%% file: main.tex
\documentclass{article}

\usepackage{microtype}
\usepackage{graphicx}
\usepackage{booktabs} 
\usepackage{subcaption} 

\usepackage{hyperref}

\usepackage[accepted]{mlsys2023}

\usepackage{float}
\usepackage{amsmath}
\usepackage{amsfonts}
\usepackage{etoolbox}
\usepackage{multirow}
\usepackage{xspace}
\usepackage{pgfplots} 
\pgfplotsset{width=13cm, compat=1.6}
\usetikzlibrary{patterns}
\usepackage{xcolor}

\include{macros}
\usepackage{enumitem}
\setlist{nosep}

\usepackage[english]{babel}
\addto\extrasenglish{

}
\usepackage[algoruled, vlined, procnumbered]{algorithm2e}

\mlsystitlerunning{\tool: Accurate ML Inference on Microcontrollers}

\begin{document}

\twocolumn[
\mlsystitle{\tool: Accurate ML Inference on Microcontrollers}



\mlsyssetsymbol{equal}{*}

\begin{mlsysauthorlist}
\mlsysauthor{Shikhar Jaiswal}{equal,to}
\mlsysauthor{Rahul Kiran Kranti Goli}{equal,goo}
\mlsysauthor{Aayan Kumar}{ed}
\mlsysauthor{Vivek Seshadri}{to}
\mlsysauthor{Rahul Sharma}{to}
\end{mlsysauthorlist}

\mlsysaffiliation{to}{Microsoft Research, India}
\mlsysaffiliation{goo}{ETH Zurich, Switzerland}
\mlsysaffiliation{ed}{UC Berkeley, USA}

\mlsyscorrespondingauthor{Shikhar Jaiswal}{jaiswalshikhar87@gmail.com}

\mlsyskeywords{TinyML,
Memory Management,
Programming Languages,
Compilers,
Number Representations,
Embedded Devices}

\vskip 0.3in

\begin{abstract}
Running machine learning inference on tiny devices, known as TinyML, is an emerging research area. This task requires generating inference code that uses memory frugally, a task that standard ML frameworks are ill-suited for. A deployment framework for TinyML must be a) parametric in the number representation to take advantage of the emerging representations like posits, b) carefully assign high-precision to a few tensors so that most tensors can be kept in low-precision while still maintaining model accuracy, and c) avoid memory fragmentation. We describe \tool, the first TinyML framework that holistically addresses these issues to generate efficient code for ARM microcontrollers (e.g., Arduino Uno, Due and STM32H747) that outperforms the prior TinyML frameworks.
\end{abstract}
]



\printAffiliationsAndNotice{\mlsysEqualContribution} 

\input{sections/1-introduction}

\input{sections/8-related-work}

\input{sections/4-compiler-overview}

\input{sections/5-technical-details}
\input{sections/6-implementation-details}

\input{sections/7-evaluation}

\input{sections/9-conclusion}

\bibliography{bibfile}
\bibliographystyle{mlsys2023}
\clearpage

\appendix

\input{sections/2-preliminaries}

\input{sections/3-working-example}

\input{sections/12-haunter-extra-details}

\input{sections/11-appendix}


\end{document}

%% file: macros.tex
\newcommand{\softposit}{\textsc{SoftPosit}\xspace}
\newcommand{\bfloath}{\textsc{BFloat}\xspace}
\newcommand{\gemmlowp}{\textsc{Gemmlowp}\xspace}
\newcommand{\bfloat}{\textsc{BFloat16}\xspace}
\newcommand{\quarterprecision}{\textsc{Fixed8}\xspace}
\newcommand{\quarterprecisionposit}{\textsc{Posit8}\xspace}
\newcommand{\halfprecision}{\textsc{Fixed16}\xspace}

\newcommand{\fullprecision}{\textsc{Float32}\xspace}

\newcommand{\tool}{\textsc{MinUn}\xspace}
\newcommand{\toolfixed}{\textsc{MinUn-Fixed}\xspace}
\newcommand{\toolposit}{\textsc{MinUn-Posit}\xspace}
\newcommand{\previoustooldsl}{\textsc{Shiftry-DSL}\xspace}
\newcommand{\previoustool}{\textsc{Shiftry}\xspace}
\newcommand{\previousprevioustool}{\textsc{SeeDot}\xspace}
\newcommand{\previoustoolfixed}{\textsc{Shiftry-Fixed}\xspace}
\newcommand{\previoustoolposit}{\textsc{Shiftry-Posit}\xspace}
\newcommand{\fastgrnn}{\textsc{FastGRNN}\xspace}
\newcommand{\protonn}{\textsc{ProtoNN}\xspace}
\newcommand{\bonsai}{\textsc{Bonsai}\xspace}
\newcommand{\rnnpool}{\textsc{RNNPool}\xspace}

\newcommand{\haunter}{\textsc{Haunter}\xspace}
\newcommand{\tflite}{\textsc{TFLite}\xspace}
\newcommand{\zskew}{\textsc{Zero-Skew8}\xspace}

\newcommand\mydots{\hbox to 1em{.\hss.\hss.}}

\mathchardef\mhyphen="2D

\newcommand{\positg}[2]{
        \ifstrempty{#2}{
            \ensuremath{\mathtt{quire}_{#1}}
        }
        {
            \ensuremath{\mathtt{posit}_{#1}{\mathtt{#2}}}
        }
}

\definecolor{bblue}{HTML}{4F81BD}
\definecolor{rred}{HTML}{C0504D}
\definecolor{ggreen}{HTML}{9BBB59}
\definecolor{ppurple}{HTML}{9F4C7C}
\definecolor{yyellow}{HTML}{DDAE06}

\tikzstyle{bar1} = [bblue, fill=bblue, postaction={pattern=north east lines}]
\tikzstyle{bar2} = [rred, fill=rred]
\tikzstyle{bar3} = [ggreen, fill=ggreen, postaction={pattern=north west lines}]
\tikzstyle{bar4} = [ppurple, fill=ppurple, postaction={pattern=crosshatch dots}]
\tikzstyle{bar5} = [yyellow, fill=yyellow, postaction={pattern=grid}]

\newcommand{\graph}[9]{
    \begin{tikzpicture}  
    \begin{axis}  
    [  
        bar width=0.1cm,
        ybar, 
        #3,
        #2,
        height=6cm,
        width=7.2cm,
        enlarge x limits=0.18,
        legend style={font=\scriptsize, at={(0.5,0)},
          anchor=north,legend columns=2},
        legend pos = north east,
        ylabel={#9}, 
        symbolic x coords={#4},  
        xtick=data,  
        xticklabel style={
                font=\scriptsize,
                text width=0.5cm,
                align=left,
            },
        ] 
    #5
    \legend{#6}  
    \ifstrempty{#7}{}
	{
	    \coordinate (A) at (axis cs:#8,#7);
        \coordinate (O1) at (rel axis cs:0,0);
        \coordinate (O2) at (rel axis cs:1,0);
        \draw [black,sharp plot,dashed] (A -| O1) -- (A -| O2);
	}
    
    \end{axis}
    \end{tikzpicture}
    
}

%% file: sections/1-introduction.tex
\section{Introduction}

Even though memory is one of the most expensive resources, widely-used machine learning (ML) frameworks like TensorFlow and PyTorch  assume the availability of plentiful memory during inference. Memory constraints are hard: exceeding the available memory by a single byte will cause a crash. Hence, these frameworks, which rely on interpreters, are unsuitable for running ML on memory-constrained devices.
For example, an Arduino Uno~\cite{uno} has $2$ KBs of RAM then no ML interpreter can fit in it.
In this paper, we explore deployment frameworks for running ML inference  on such devices.

Recently, there has been a flurry of work in the TinyML\footnote{\url{https://www.tinyml.org/}} space that aims to run ML on low-power embedded devices~\cite{mafia, shiftry, seedot, protonn, bonsai, fastgrnn, emirnn, sharnn, rnnpool, haq, mcunet, mcunetv2}.
As a motivating example, consider a low-power chip embedded in the brain that can detect the onset of seizures and warn the user~\cite{bonsai}. Animal testing of brain chips is already under way by {\sc NeuraLink}~\cite{neuralink}. Such chips need to both perform inference locally on the device (user might be in a place with no internet connectivity) and be low-power (to avoid tissue damage caused by heat dissipation and brain surgeries for battery replacement).
Since maintaining large RAMs drains batteries, these tiny devices need to have small RAMs, ranging from a few bytes to a few kilobytes (KBs). Although, the technical specifications of {\sc NeuraLink} are unavailable in public domain, for reference, the microcontrollers in pacemakers in hearts have memories between 16 bytes and $10$ KBs~\cite{pacemakermem, pacemaker}. Hence, there is a need for deployment frameworks that compile ML models to memory efficient code that can run within the specified RAM limits of the tiny devices.

\begin{figure}
    \centering
    \includegraphics[scale=0.7]{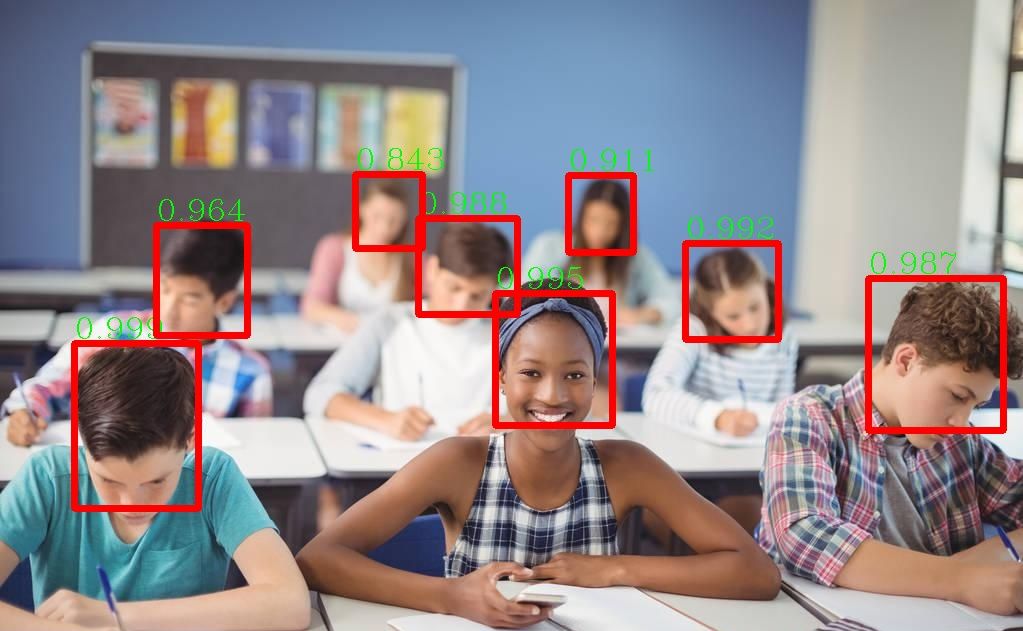}
    \caption{Face detection example.}
    \label{fig:head}
    \vspace{-10pt}
\end{figure}

Running ML inference on tiny devices is non-trivial.
For example, Convolutional Neural Networks (CNNs) have millions of parameters,  require MBs or GBs of memory, and are unsuitable to be run on tiny devices. Hence, we are interested in recent models targeted for tiny devices that have been specifically designed to provide good accuracy while requiring only thousands of parameters and  KBs of memory~\cite{protonn, bonsai, fastgrnn, rnnpool}, e.g., Recurrent Neural Networks (RNNs).
However, even running these models on tiny devices is challenging.
All such ML models use high-precision 32-bit floating-points and we want to squeeze them onto tiny devices by further reducing bitwidths.
For example, suppose we want to run {\em face detection}~\cite{extd,lffd,faceboxes,eagleeye}  on a standard ARM Cortex-M class microcontroller with $256$ KBs of RAM~\cite{stm}. See \textbf{\autoref{fig:head}} for an  example where an ML model has marked the heads in the image. 
The \rnnpool~\cite{rnnpool} model provides state-of-the-art accuracy while consuming only $\sim 600$ KBs of RAM, which is a huge improvement over prior models, for e.g., even size-optimized architectures like MobileNetV2-SSDLite need over $3$ MBs of RAM. However, this \rnnpool model still exceeds the  available memory of the 256 KB device. 

An approach that has been well-studied in the literature is to compile or retrain 32-bit floating-point ML models to low bitwidth 8-bit or 2-bit or 1-bit code~\cite{tflite, dfq, krishnamoorthi, meller, metaQuant, cnnQuant, lstmQuant, quantNetTrain, differentiableQuant, ternaryNN, lossAwareQuant, relaxedQuant, quantBackProp, multiCodebookQuant}. However, the bitwidth required to maintain accuracy depends on the number of model parameters. In particular, it is well-known that models with fewer parameters need larger bitwidths~\cite{matthai} to maintain accuracy. The techniques that use bitwidths $\leq 8$
have only been shown to succeed on large models with millions of parameters.
For the models of our interest, e.g., the \rnnpool-based face detection model, even 8-bits are insufficient to maintain model accuracy (\textbf{\autoref{sec:8bitAccuracy}}). 

In the TinyML space, Tensorflow-Lite (\tflite)~\cite{tflite} converts floating-point CNNs to models that uses 8-bits for weights and 32-bits for biases; these models are then run in the \tflite interpreter that has its own memory overheads. 
\previousprevioustool~\cite{seedot} is more expressive and compiles arbitrary models to 16-bit fixed-point C++ code that runs on bare metal.
\previoustool~\cite{shiftry} goes a step further and compiles to a mixture of 8-bit and 16-bit fixed-point C++ code.
Our evaluation shows that these prior frameworks 
are unsatisfactory when addressing the three primary subproblems of TinyML.

\subsection{The Three Subproblems of TinyML}
First, we need to use number representations that can both approximate the 32-bit floating-point numbers well with a smaller number of bits, and operations on which are efficiently implementable in hardware. 
Recently there has been a wave of new number representations~\cite{fbfp,fxpnet,shiftcnn,flexpoint,ldr,inq}: \tflite's {\em Zero-Skew}, Google's {\em BFloat16}, NVIDIA's {\em TensorFloat-32}, Microsoft's {\em MSFP}, Tesla's {\em CFP8} and the {\em posit} \cite{gustafson1, gustafson2} representation. Posits are attractive as they have better dynamic range and precision compared to floating-point. Prior frameworks for TinyML are tied to specific representations like zero-skew~\cite{tflite} or fixed-point~\cite{seedot, shiftry}. A robust framework that remains relevant with rapidly evolving number representations must be parametric in the number representation. 

Second, for each tensor\footnote{In TinyML, it is common for all elements of a tensor to have the same bitwidth to avoid the memory overheads of  book-keeping the bitwidths associated with the individual elements.}, we need to decide whether to keep it in high precision or in low precision. Keeping all tensors in low precision leads to huge accuracy loss while keeping them all in high precision is wasteful memory-wise (\textbf{\autoref{sec:evaluation}}).
Hence, for $N$ tensors, we have $2^N$ choices and need a heuristic to select a good {\em bitwidth assignment} that minimizes memory, while retaining model accuracy. Crucially, the decision of whether a tensor is assigned low-precision or high-precision must take into account {\em both} the size of the tensor, and the impact it has on the precision. No prior work provides such a technique to obtain a good bitwidth assignment. Note that \tflite and \previousprevioustool use the same bitwidth assignment for all models, whereas, \previoustool exclusively uses accuracy to find a suitable bitwidth assignment.

Third, unlike modern CPUs, tiny devices have no hardware support for virtual memory. Therefore, {\em fragmentation} can quickly make a program go out of memory. Whether a program can fit in a given memory limit or not is an NP-complete bin packing problem. The solutions provided by prior works for this {\em memory management} problem, i.e., deciding the physical memory address at which a new tensor is allocated, are unsatisfactory: \tflite asks programmers to manually\footnote{\url{https://www.tensorflow.org/lite/microcontrollers\#limitations}} manage  RAM, \previousprevioustool uses wasteful static allocation, and \previoustool uses heuristics  to reduce fragmentation that have no optimality guarantees.
Although reusing memory for different variables has the flavor of {\em register allocation}~\cite{dragonbook}, note that registers don't suffer from fragmentation, which is the primary problem here.

\subsection{Our Contributions}
We provide a framework, \tool\footnote{Code and datasets available at: \url{https://github.com/krantikiran68/EdgeML/tree/shikhar_posit_vbw_haunter}}, that makes significant advances on all three subproblems. \tool is the first TinyML framework which is parametric over any arbitrary number representation and we evaluate \tool with both fixed-point and posit representations. 
We have designed \tool to have a clear separation between the representation-specific and the representation-independent parts (\textbf{\autoref{sec:bitwidth}}). In contrast, the prior TinyML frameworks are extremely tied to the number representation that they work with, e.g., fixed-point for \previousprevioustool and \previoustool.


For the bitwidth assignment problem, we propose a novel exploration algorithm, \haunter (\textbf{\autoref{sec:bitwidth}}), which uses both accuracy and size to produces better assignments than the accuracy-based heuristic of \previoustool, while being much more efficient and scalable (\textbf{\autoref{sec:complexity}}).

Finally, for RAM management, \tool encodes the memory management problem to a bin-packing problem and solves it using Knuth's Algorithm X~\cite{algoX} (\textbf{\autoref{sec:DLX}}), which is guaranteed to return the optimum result albeit in exponential time. Here, our main contribution is to come up with an effective encoding and to adapt the general framework of Algorithm X to ensure a tractable runtime in practice.  
In particular, for the model in \textbf{\autoref{fig:head}}, \tool reduces the RAM consumption from $\sim600$ KBs to $\sim180$ KBs, thus enabling this model to run on devices with $256$ KBs of RAM. Although \tool was not designed for large models, even on an ImageNet-scale CNN~\cite{squeezenet}, \tool outperforms the prior TinyML frameworks (\textbf{\autoref{sec:squeezenet}}). 

We believe that \tool has an important role in the repertoire of techniques to reduce memory usage of ML models. We anticipate that the ML researchers will apply various techniques like structural pruning, low-rank approximation of weight matrices, novel model architectures, etc., to create models with fewer parameters and \tool will further reduce the memory usage by keeping most of these parameters in low bitwidths. 
For example, the models used in our evaluation are based on novel ML architectures that keep the number of model parameters small, and \tool further reduces the peak RAM usage by keeping most of the model parameters in lower bitwidth. Hence, the ``post training quantization'' of \tool is complementary to the quantization techniques used during training.

Next, we discuss related work (\textbf{\autoref{sec:relatedWork}}) followed by an overview of \tool (\textbf{\autoref{sec:compilerOverview}}). \textbf{\autoref{sec:technical}} contains technical details, \textbf{\autoref{sec:implementation}} contains implementation details and \textbf{\autoref{sec:evaluation}} presents our empirical results.
Additional background is provided in \textbf{\autoref{sec:prelims}}
and a worked out example in \textbf{\autoref{sec:workingExample}}.







%% file: sections/8-related-work.tex
\section{Related Work}
\label{sec:relatedWork}

Running ML models with minimum resources is a vast subject and we do not attempt to survey it here. The design of new CNN architectures manually~\cite{densenet,mobilenets} or automatically with Network Architecture Search (NAS)~\cite{proxylessnas,efficientnets,squeezenas} focuses on reducing model size and compute requirements but not the peak RAM usage, which is our focus here; if the peak RAM usage exceeds the available RAM then the model fails. Techniques like pruning channels/filters~\cite{channel} and spatial down-sampling~\cite{rnnpool} reduce peak memory usage and are complementary to \tool.

There are various approaches for quantization in the ML literature. In {\em hybrid-quantization}~\cite{squeezenet,tflite} on smartphones, each low bitwidth tensor is converted to 32-bit floating-point at runtime, computed upon in floating-point, and then converted back to low bitwidth. This approach is untenable for the models and devices that we consider because  we have observed that even converting a single quantized tensor to 32-bit  overflows the RAM.
Apart from \previoustool~\cite{shiftry}, prior papers on quantization use the trivial bitwidth assignment of mapping all tensors in large CNNs to low bitwidths~\cite{binary1, binary2, related7, related8, metaQuant, cnnQuant, lstmQuant, quantNetTrain, differentiableQuant, ternaryNN, lossAwareQuant, relaxedQuant, quantBackProp, multiCodebookQuant, dfq, krishnamoorthi, meller, tflite}. Most of these approaches rely on retraining a low-bitwidth model, which is a computationally very expensive process. Thus, ``post-training quantization'' frameworks~\cite{tflite, seedot, shiftry} like \tool offer a distinct advantage.

\tool is related to floating-point tuning algorithms like Precimonious~\cite{precimonious}. However, their objective is completely orthogonal to the problem considered by \tool. \tool aims to maximize the accuracy of a program while strictly respecting the memory usage constraints, while the Precimonious algorithm aims to maximize performance (in terms of latency of execution) while strictly respecting the precision constraints. We are not aware of any floating-point heuristic tuning algorithm which has the same objectives as \tool.

Memory management in the space of embedded devices is currently performed via wasteful static allocation, fragmentation-prone dynamic allocation, memory pools with restricted applicability, and so on.
~\citet{opt-stack} use Integer Linear Programming (ILP) to generate optimal allocation schemes for a memory management problem that is different from ours and their allocations are suboptimal for the problem we consider.

\tool can be thought of as an approximation framework for ML models.
Most existing approximation frameworks  map floating-point programs to other floating-point programs~\cite{douskos,precimonious,zhu,stokefp,green,herbie}. Converters from floating-point to fixed-point that have not been designed for ML include~\cite{rosa,darulova1,darulova2,rinard,brooks,haldar,banerjee,bevcvavr,menard,fridge}. 

%

%% file: sections/4-compiler-overview.tex
\section{Compiler Overview}
\label{sec:compilerOverview}
\begin{figure}[ht]
    \centering
    \includegraphics[scale=0.29]{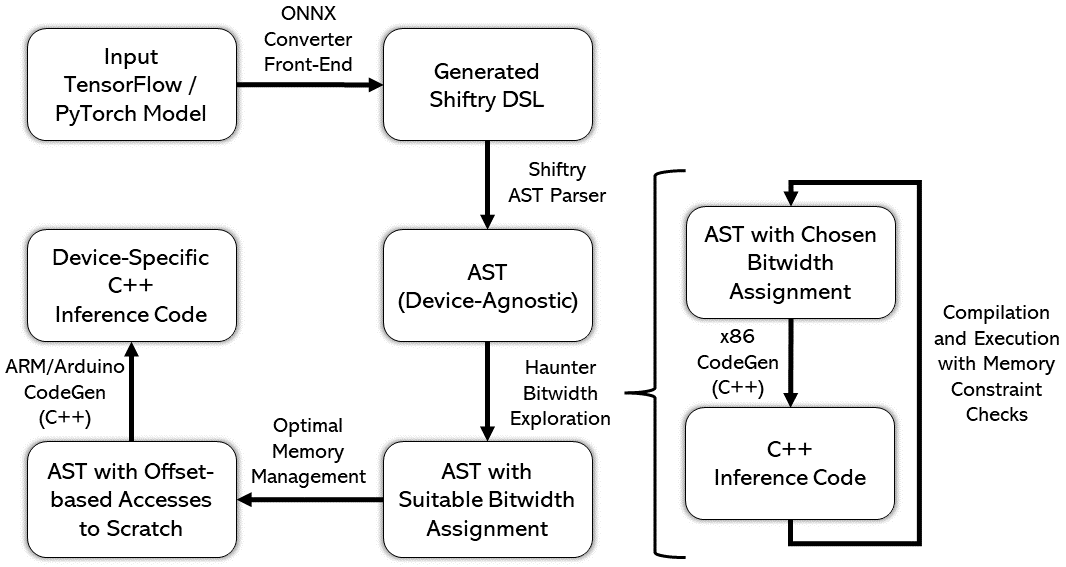}
    \caption{\tool Overview.}
    \label{fig:overview}
    \vspace{-10pt}
\end{figure}

In this section, we provide a high-level overview of the \tool compiler (\textbf{\autoref{fig:overview}}). \tool expects as input a program written in the \previoustooldsl~\cite{shiftry}, a concise syntax for expressing operations and functions commonly used in ML models. 

The DSL code for models can either be written manually or automatically obtained by exporting TensorFlow/PyTorch models to ONNX and using \tool's ONNX front-end (\textbf{\autoref{sec:implementation}}). 
\tool then parses the input program from the DSL and generates an abstract syntax tree (AST) required for \haunter, our bitwidth exploration strategy. \haunter then returns a suitable bitwidth assignment for all the tensors in the program. This is achieved by choosing an initial bitwidth assignment, generating the corresponding x86 C++ inference code for it, compiling it using GCC and executing the code for logging accuracy and memory usage, and using the logged information for making changes to the bitwidth assignment. \haunter ensures that user-provided memory constraints are always met during bitwidth exploration, and returns a suitable assignment which maximizes accuracy.

Then through our optimum memory management strategy, we obtain a {\em memory map} from RAM tensors to indices in a global {\em scratch} buffer, corresponding to the given bitwidth configuration. Recall that the aim of \tool is to minimize the total size of {\em scratch} while ensuring that the program executes correctly.

The final device-specific C++ code is generated using a codegen pass over the AST, which also replaces all the intermediate RAM tensors, with offsets-based accesses to  {\em scratch} on the basis of our memory map. This final C++ code is then cross-compiled (using either ARM GNU Toolchain or Arduino IDE) and executed on the microcontroller of choice. Further, we exposit the technical details, and analyze the empirical benefits of \tool.

%% file: sections/5-technical-details.tex
\section{Technical Details}
\label{sec:technical}
In this section, we explain \haunter, analyze its complexity, and describe our  memory management mechanism.
A working example is provided in \textbf{\autoref{sec:workingExample}} and 
detailed algorithms used by \haunter are provided in \textbf{\autoref{sec:haunter-details}}.

\subsection{Bitwidth Assignment with \haunter}
\label{sec:bitwidth}
The \haunter algorithm works in a three-stage process, with the first stage being dependent on the number representation being considered, and the remaining two stages staying agnostic to the representation. \haunter uses three pieces of user-provided information - the amount of available RAM, a pair of bitwidths (\textit{highBitwidth} and \textit{lowBitwidth}) between which the exploration is to be commenced, and a soft limit factor that acts as a tuning parameter
 which can be used to make the exploration process more frugal, by choosing a soft limit $<1$. We use $\rho$ to denote the bitwidth assignment, that maps each tensor to two possible bitwidths \{\textit{lowBitwidth}, \textit{highBitwidth}\}. Note that while we limit most discussions to 8-bit / 16-bit to simplify exposition and allow comparison, \haunter 
generalizes in a straightforward manner to exploration across various bitwidths
(\textbf{\autoref{sec:choice}}).

\subsubsection{\textit{Preprocessing (Stage I)}} This stage finds the values of data-dependent parameters  that capture the runtime ranges of inputs and intermediate tensors. Examples of such parameters include the most suitable value of \textit{es} (\textbf{\autoref{sec:positprelims}}) for 8-bit and 16-bit posits (since multiple \textit{es} options are supported for these bitwidths), and scales (\textbf{\autoref{sec:fixedprelims}}) for individual tensors for fixed-point. \textbf{\autoref{alg:preprocessing}} in \textbf{\autoref{sec:haunter-details}} succinctly describes the driver method used for obtaining these parameters for posits, where we simply pick an appropriate \textit{es} value based on accuracy obtained with homogeneous bitwidth configurations. To determine scales of fixed-point, we use \previoustool's data-driven scale assignment procedure~\cite{shiftry}, that profiles each tensor in the floating-point version of the code by running it on a given set of inputs. This profiling information includes the minimum and maximum values observed for each tensor, and is used for assigning a suitable scale to the fixed-point tensor (\textbf{\autoref{sec:fixedprelims}}).

\subsubsection{\textit{Heat-Map Generation (Stage II)}}
\label{sec:heatMapGen}
During inference on a particular data point, a ``value map'' stores key-value pairs, where the keys are the tensor names, and the values are the floating point values held by that tensor during execution on the provided data point. \haunter starts from an initial configuration where all the tensors are kept in \textit{lowBitwidth}. In the first phase (\textbf{\autoref{alg:valMaps}} in \textbf{\autoref{sec:haunter-details}}), we set each promotable tensor to \textit{highBitwidth} simultaneously, and then compile and execute the code to obtain a ``high-bitwidth value map''. We repeat the same procedure once more, by simultaneously setting each tensor to \textit{lowBitwidth}, compiling and executing to obtain a similar ``low-bitwidth value map''.

Based on these two maps, we attempt to create a ``heat map'' in the second phase (\textbf{\autoref{alg:heatMap}} in \textbf{\autoref{sec:haunter-details}}), wherein, we simply calculate the absolute deviation in the representative floating-point values between the \textit{highBitwidth} and \textit{lowBitwidth} case for each tensor. Since keeping track of large tensors is infeasible, we simply sort the element-wise deviations in increasing order and take the 95th percentile deviation as the representative error deviation for each tensor. We define the promotability score as:
$$ \text{Promotability} = \frac{\text{95th Percentile Error Deviation}}{\text{Tensor Cardinality}}$$
Here, tensor cardinality  refers to the size of the tensor or the number of elements in a tensor. Sorting the tensors in decreasing order of the promotability scores provides an ordered ranking called \textit{promotionOrder}, which prioritizes smaller-sized tensors with large error deviations to be promoted first.
One could advocate the use of accuracy values as the metric to be used in place of error deviation for promotability scores. While it is a more direct metric for promoting tensors, this strategy, like \previoustool's bitwidth exploration, suffers from higher computational complexity ({\bf \autoref{sec:complexity}}).

\subsubsection{\textit{Promotion Algorithm (Stage III)}}
\label{sec:promotionAlgo}
The promotion algorithm (\textbf{\autoref{alg:attackingAlgo}} in \textbf{\autoref{sec:haunter-details}}) itself runs in three stages. The first stage aims to cumulatively promote all the tensors while staying within the \textit{memoryLimit}, and also determine which tensors can potentially overshoot the provided limit. The second stage individually considers these ``overshooting'' tensors, by promoting them first, and then commencing the \textit{promotionOrder}-based exploration. The final stage commences the exploration after simultaneously promoting all of these ``overshooting'' tensors. The final bitwidth configuration is chosen as the one minimizing the disagreement counts against the floating point code outputs, and is returned as the output of the bitwidth exploration.

\subsection{Complexity Analysis}
\label{sec:complexity}
Bitwidth exploration, via \haunter, is the most computationally expensive block of \tool, owing to repeated code code generation and compilation for varying bitwidth assignments, and execution calls for measuring the accuracy. While the former depends on the size of the program text of the model, the latter is determined by both the runtime of the model and the size of the dataset being considered.

We denote the amount of time required per code generation call as $\mathcal{T_{\mathit{codegen}}}$, and treat it as the upper bound of the code generation and compilation time required for different bitwidth configurations. We follow the same terminology for execution call latency (per example) as well, terming it $\mathcal{T_{\mathit{execution}}}$.

For a model of $N$ tensors, supplied with a dataset of $D$ samples, \previoustool incurs a bitwidth exploration latency of:
$$\mathcal{O}(N \times D \times \mathcal{T_{\mathit{execution}}}+N \times \mathcal{T_{\mathit{codegen}}} )$$
which \haunter improves to (see \textbf{\autoref{sec:complexity-details}}):
$$\mathcal{O}(\log{}N \times D \times \mathcal{T_{\mathit{execution}}}+N \times \log{}N \times \mathcal{T_{\mathit{codegen}}} )$$

\textbf{\autoref{tab:fixedPointTimesBrief}} and \textbf{\autoref{tab:fixedPointTimes}} in \textbf{\autoref{sec:appendix}} reports the empirical reduction in code generation and compilation time.

\subsection{Optimum Memory Management}
\label{sec:DLX}

\begin{figure*}
\centering
\begin{minipage}{\columnwidth}
  \centering
    \input{figures/fragmentation}
\end{minipage}
\hfill
\begin{minipage}{\columnwidth}
  \centering
    \input{figures/bin_packing}
\end{minipage}
\vspace{-10pt}
\end{figure*}

Once the bitwidth assignment has been computed, \tool must allocate tensors at physical memory addresses to keep the peak RAM consumption low by maximizing the reuse of RAM locations. There are several ways to do this memory management.

\textit{Static Assignment}: Tensors are pushed onto the program stack when they come in scope and popped out when they go out of scope in Last-In-First-Out (LIFO) order. This approach results in high RAM consumption as a tensor $X$ cannot be deallocated before tensors instantiated later are deallocated, even when $X$ is no longer live.

\textit{Dynamic Assignment}: Tensors are allocated memory on the program heap, and freed when no longer needed. This approach delegates the memory management to the runtime, which may not be preferable for resource-constrained devices owing to their runtime overheads. This approach is known to be prone to fragmentation~\cite{shiftry}, as depicted in \textbf{\autoref{fig:fragmentation}}.
Here, deallocating variables $A$ and $C$ has fragmented the memory.

\textit{Delegating Memory Management to the Compiler}: Given an ML model with a fixed-sized input, tensor sizes and live ranges (range of instructions where a tensor is used and needs memory) can be known at compile time. A mapping from tensors to memory locations (or indices) on a contiguous array ({\em scratch}) can be computed during the code generation process itself. This reuses the memory for tensors that are no longer alive to allocate new tensors.  For example, in \textbf{\autoref{fig:fragmentation}}, one way to avoid fragmentation is for the compiler to map $A$ and $C$ to memory locations that are adjacent to each other. Computing such maps is an NP-Hard bin packing problem which can be addressed through suboptimal greedy heuristics. 

Static and Dynamic Assignment methods fail to fit \fastgrnn and \rnnpool models for the typical RAM budgets that are available with microcontrollers. Hence, we take the third approach, but unlike prior work~\cite{shiftry} we do not use greedy heuristics. Instead, we approach the bin-packing problem as an exact cover problem.

\textbf{Definition 1 (Exact Cover)}: Let $\mathcal{P}(X)$ denote the power set of a set $X$. Given a set $X$ and a set $S \subset \mathcal{P}(X)$, an exact cover of $X$ is a set $S^{*} \subset S$ such that:
\begin{itemize}
    \item $s_{i} \cap s_{j} = \emptyset, \forall s_{i}, s_{j} \in S^{*},  i \neq j$
    \item $\bigcup\limits_{s_{i} \in S^{*}} s_{i} = X$
\end{itemize}


Consider a 2-D canvas of size $M \times I$ (where $M$ is the memory budget in bytes along y-axis and $I$ is the number of instructions in the program along x-axis). The smallest individual element $c_{i}$ of the canvas is a square of unit area, semantically denoting a byte-sized tensor which is active only for a single instruction.
For a tensor $v_{i}$ active between instructions $i_{\mathit{start}}$ and $i_{\mathit{end}}$, and occupying $b_i$ bytes in memory, we can map a rectangle made of $b_i \times (i_{\mathit{end}} - i_{\mathit{start}})$ contiguous units on this canvas, occurring between $i_{\mathit{start}}$ and $i_{\mathit{start}}$.

Let $C = \{c_{1}, c_{2}, \dots, c_{M\times I}\}$ denote the set of all unit elements of the canvas, and $V = \{v_{1}, v_{2}, \dots, v_{N}\}$ denote the set of all tensors in the program. Additionally, for each tensor $v_{i}$, we define a new set $C_{v_{i}} \subset \mathcal{P(C)}$ containing all subsets of unit elements in $C$, which can be potential memory maps to the tensor. Hence, each individual set in $C_{v_{i}}$ forms a contiguous block of memory spanning the live range of $v_{i}$, and has a height of at least $b_{i}$ units, as depicted in \textbf{\autoref{fig:binPacking}}.

Let $X = C \cup V$ and  $S \subset \mathcal{P}(X)$ with the following constraint:
$\forall s_{k} \in S, v_{i} \in V [v_{i} \in s_{k} \Rightarrow s_{k} \setminus v_{i} \in C_{v_{i}}]$

Given these sets, we solve the exact cover problem using an exponential-time, backtracking search algorithm called Algorithm X~\cite{algoX}, with the following optimizations to speed up the exploration:
\begin{itemize}
    \item We use the Dancing Links Algorithm as proposed in~\cite{algoX} to efficiently assign and unassign a tensor rectangle to a particular memory location.
    \item We visit the tensors in decreasing order of the area of the rectangles, as a heuristic to quickly prune out assignments causing conflicts between tensors.
    \item We make a coarsening assumption: all tensors are assumed to have a size as a multiple of some coarsening constant $k$. For example, if a program has $3$ tensors of size $32$, $56$ and $64$ bytes, then we can set $k = 32$, rounding up the sizes of the tensors as needed. Once all tensor sizes are a multiple of $k$, we can effectively create a canvas of size $\frac{M}{k} \times I$ instead of $M \times I$ by dividing all sizes by $k$, which reduces the problem size. Note that choosing $k = 1$ recovers the original formulation of the algorithm, and will always find the optimal solution, if it exists.
\end{itemize}

Our optimum memory management mechanism aims to find an exact cover for the minimum viable value of $M$. At each instruction, we can find minimum memory budget required for execution by tightly stacking the alive tensors at that instruction, with no gaps and no overlaps along the y-axis, and looking at height of the stack. Calculating this height for all $I$ instructions, and picking the highest value out of them, gives us this minimum budget.

Once an exact cover is found for the minimum $M$, we can map a tensor from the canvas to the {\em scratch}, by taking its minimum y-coordinate, and using that as an offset from the starting location of the {\em scratch}. Notice that since $M$ denotes the minimum viable height of the canvas, it also denotes the total size of the {\em scratch}. Since exact cover finds a suitable allocation within the minimum viable budget, it effectively resolves the fragmentation problem.

The proof regarding the optimality of our memory management technique reduces to the proof that Algorithm X will always find an optimum assignment if it exists, which is well-known.

%% file: figures/fragmentation.tex
\begin{figure}[H]
    \centering
    \begin{tikzpicture}[domain=0:2, scale=0.7, every node/.style={scale=0.7}]
        \draw[thick,color=gray,step=1cm] (0,0) grid (6,1);
        \draw[->] (0,0) -- (6.5,0)
        node[below right] {$\mathit{RAM}$};
        \foreach \x in {0,64,128,192,256,320, 384}
            \node at (\x/64,-7pt) {$\x$};
    \end{tikzpicture}
    \begin{tikzpicture}[domain=0:2, scale=0.7, every node/.style={scale=0.7}]
        \draw[thick,color=gray,step=1cm] (0,0) grid (6,1);
        \draw[thin,color=red,step=1cm,fill=red!10,] (0,0) grid (1,1) rectangle (0,0);
        \node at (0.5,0.5) {$A$};
        \draw[thin,color=blue,step=1cm,fill=blue!10,] (1,0) grid (2,1) rectangle (1,0);
        \node at (1.5,0.5) {$B$};
        \draw[thin,color=orange,step=1cm,fill=orange!10,] (2,0) grid (3,1) rectangle (2,0);
        \node at (2.5,0.5) {$C$};
        \draw[thin,color=black,step=1cm,fill=black!10,] (3,0) grid (4,1) rectangle (3,0);
        \node at (3.5,0.5) {$D$};
        \draw[->] (0,0) -- (6.5,0)
        node[below right] {$\mathit{RAM}$};
        \foreach \x in {0,64,128,192,256,320, 384}
            \node at (\x/64,-7pt) {$\x$};
    \end{tikzpicture}
    \begin{tikzpicture}[domain=0:2, scale=0.7, every node/.style={scale=0.7}]
        \draw[thick,color=gray,step=1cm] (0,0) grid (6,1);
        \draw[thin,color=blue,step=1cm,fill=blue!10,] (1,0) grid (2,1) rectangle (1,0);
        \node at (1.5,0.5) {$B$};
        \draw[thin,color=black,step=1cm,fill=black!10,] (3,0) grid (4,1) rectangle (3,0);
        \node at (3.5,0.5) {$D$};
        \draw[->] (0,0) -- (6.5,0)
        node[below right] {$\mathit{RAM}$};
        \foreach \x in {0,64,128,192,256,320, 384}
            \node at (\x/64,-7pt) {$\x$};
    \end{tikzpicture}
    \begin{tikzpicture}[domain=0:2, scale=0.7, every node/.style={scale=0.7}]
        \draw[thick,color=gray,step=1cm] (0,0) grid (6,1);
        \draw[thin,color=blue,step=1cm,fill=blue!10,] (1,0) grid (2,1) rectangle (1,0);
        \node at (1.5,0.5) {$B$};
        \draw[thin,color=black,step=1cm,fill=black!10,] (3,0) grid (4,1) rectangle (3,0);
        \node at (3.5,0.5) {$D$};
        \draw[thin,color=violet,step=1cm,fill=violet!10,] (4,0) grid (6,1) rectangle (4,0);
        \node at (5.0,0.5) {$E$};
        \draw[->] (0,0) -- (6.5,0)
        node[below right] {$\mathit{RAM}$};
        \foreach \x in {0,64,128,192,256,320, 384}
            \node at (\x/64,-7pt) {$\x$};
    \end{tikzpicture}
    \caption{Fragmentation example: 1) Start with an empty RAM. 2) Allocate tensors $A$, $B$, $C$, and $D$, each of $64$-bytes. 3) Later, deallocate $A$ and $C$, freeing $128$ bytes. 4) Then allocate $E$ of size $128$ bytes. Fragmented RAM Usage: 384 bytes, Optimal Peak RAM Usage: 256 bytes.}
    \label{fig:fragmentation}
\end{figure}
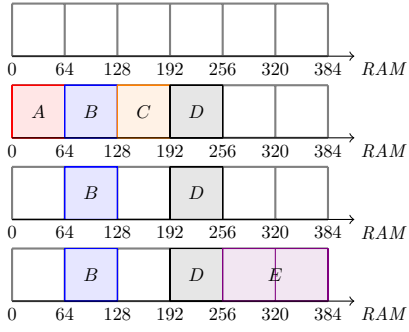

%% file: figures/bin_packing.tex
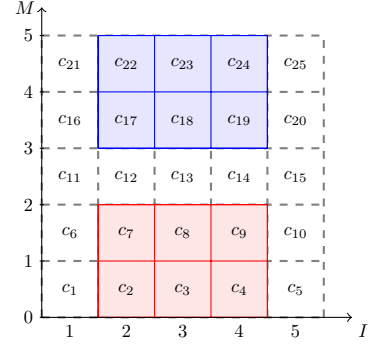
\begin{figure}[H]
    \centering
    \begin{tikzpicture}[domain=0:2, scale=0.75, every node/.style={scale=0.7}]
        \draw[thick,color=gray,step=1cm,
        dashed] (0,0) grid (5,5);
        \draw[thin,color=red,step=1cm,fill=red!10,] (1,0) grid (4,2) rectangle (1,0);
        \draw[thin,color=blue,step=1cm,fill=blue!10,] (1,3) grid (4,5) rectangle (1,3);
        \foreach \y in {4,...,0}
            \foreach \x in {1,...,5}
                \pgfmathsetmacro\z{int(\x+(5*\y))}
                    \node at (\x-0.5,\y+0.5) {$c_{\z}$};
        \draw[->] (0,0) -- (5.5,0)
        node[below right] {$I$};
        \draw[->] (0,0) -- (0,5.5)
        node[left] {$M$};
        \foreach \y in {0,...,5}
            \draw (1pt,\y cm) -- (-1pt,\y cm) node[anchor=east] {$\y$};
        \foreach \x in {1,...,5}
            \node at (\x-0.5,-7pt) {$\x$};
    \end{tikzpicture}
    \caption{Example canvas with potential memory maps (in color) for a 2-byte tensor active between instructions 2 and 4. Note that each such map would form a unique element in $C_{v_{i}}$ for a given $v_{i}$.}
    \label{fig:binPacking}
\end{figure}

%% file: sections/6-implementation-details.tex
\section{Implementation Details}
\label{sec:implementation}
Our framework is built on top of \previoustool, and has been implemented in 19K lines of Python code and 12K lines of C/C++ code. Additionally, we make use of the following set of libraries in our experiments in order to obtain arithmetic routines for different number representations:

\begin{itemize}
    \item \bfloath~\cite{bfloatcc} is a standalone C++ header-only sub-library implemented in the TensorFlow library, allowing conversion between \fullprecision and \bfloat types.
    \item \softposit~\cite{softposit} provides posit arithmetic routines for simulating on x86-based platforms using C++. The library contains all functions expected in the Posit Standard~\cite{posit-standard}. For 8-bit posits, $es = 0$ and $es = 2$ support, and for 16-bit posits, $es = 1$ and $es = 2$ support is available. For 9-bit, 10-bit, 12-bit and 32-bit posits, only $es = 2$ support is available at the time of writing. 
    \item \gemmlowp~\cite{gemmlowp} is a matrix multiplication library, designed for use with ``low precision'' 8-bit fixed-point types. It is the library used by \tflite's zero-skew number representation, written in C++.
\end{itemize}


For the evaluation on large models (\textbf{\autoref{sec:squeezenet}}), we downloaded the pre-trained ONNX model from the public ONNX Model Zoo~\cite{onnx-github}, and implemented a Python front-end for automatically compiling ONNX models to the DSL (\textbf{\autoref{sec:compilerOverview}}) programs that \tool takes as input. 
In cases where certain arithmetic functions are not available for a given number representation (namely, \bfloat arithmetic, and \textit{exp(x)} function in \softposit), we simply convert the given representation to \fullprecision, fall back on the internal \fullprecision arithmetic for computation, and convert the generated output to the original type. This ensures fairness of comparison against the \fullprecision gold standard. The timeout for memory management is set as 2 hours in our evaluation. For \rnnpool models, we use a coarsening parameter of $k = 64$, and for remaining models, we use a coarsening parameter of $k = 1$.

%% file: sections/7-evaluation.tex
\section{Evaluation}
\label{sec:evaluation}
\begin{figure*}
\centering
\begin{minipage}{\columnwidth}
  \centering
  \input{figures/uniform_8_bit_accuracy}
\end{minipage}
\hfill
\begin{minipage}{\columnwidth}
  \centering
    \input{figures/float_posit_ram}
\end{minipage}
\vspace{-10pt}
\end{figure*}

We refer to \tool parameterized with the posit representation and the fixed-point representation as \toolposit and \toolfixed, respectively. We will also refer to vanilla \previoustool as \previoustoolfixed, and our adaption of \previoustool to posits as \previoustoolposit. Here, we show evaluation to justify the following claims:
\begin{enumerate}
    \item Running ML models with low-precision 8-bit representations incurs unacceptable accuracy drops (\textbf{\autoref{fig:AccuracyDrop}}). Note that high-precision 16-bit representations preserve accuracy (\textbf{\autoref{tab:floatFixedPositAccuracy}} in \textbf{\autoref{sec:appendix}}).
    \item \toolposit generated code that mixes 16-bit high-precision and 8-bit low-precision matches the accuracy of 32-bit floating-point models while consuming significantly less RAM (\textbf{\autoref{fig:RAMCompression}}).
    \item \haunter, the bitwidth assignment mechanism of \tool, outperforms the mechanisms that are exclusively accuracy-based or exclusively size-based on both fixed-point (\textbf{\autoref{tab:fixedAccuracyBrief}}) and posit representations (\textbf{\autoref{tab:positAccuracyBrief}}). 
    \item \tool-generated fixed-point code runs on real microcontrollers (\textbf{\autoref{sec:times}}). 
    \item For some models, using optimum memory management  significantly lowers the RAM consumption compared to a standard first-fit heuristic (\textbf{\autoref{sec:squeezenet}}, and  \textbf{\autoref{tab:fixedAccuracy}} and \textbf{\autoref{tab:positAccuracy}} in \textbf{\autoref{sec:appendix}}).
\item Even on an ImageNet-scale model~\cite{squeezenet}, \tool outperforms the prior TinyML frameworks (\textbf{\autoref{tab:squeezenet}}).
\end{enumerate}

We compare \tool with \previoustool~\cite{shiftry} that uses both 8-bit and 16-bit fixed-point arithmetic and \tflite \cite{tflite} that uses 8-bit zero-skew representation. Since \previousprevioustool~\cite{seedot} uses 16-bit fixed-point, it is guaranteed to have worse RAM consumption than \tool and we omit comparisons with it.

\subsection{Classification Models and Datasets}
We evaluate on three state-of-the-art TinyML classification models developed for resource constrained devices, namely \protonn~\cite{protonn}, \bonsai~\cite{bonsai}, and \fastgrnn~\cite{fastgrnn}.
These are suitable for running on Arduino class devices, which support a minimum of $2$ KBs of SRAM and $40$ KBs of Flash.
For fairness of evaluation, we cover the entire suite of datasets evaluated by \previoustool. These include CIFAR~\cite{cifar}, Character Recognition (CR)~\cite{cr}, USPS~\cite{usps}, Curet~\cite{curet}, Letter~\cite{letter}, Ward~\cite{ward} and MNIST~\cite{mnist} for \protonn and \bonsai, and DSA~\cite{dsa}, Google~\cite{google}, HAR~\cite{har}, MNIST~\cite{mnist}, USPS~\cite{usps}, Industrial~\cite{shiftry} and Wakeword~\cite{fastgrnn} for \fastgrnn.

\subsection{Localization Models and Datasets}
\label{sec:faceDetect}
\rnnpool~\cite{rnnpool} is a new pooling layer developed for reducing the sizes of convolutional outputs, while retaining sufficient information for downstream tasks, thereby saving compute and reducing memory footprint. Using \rnnpool, we design three face detection models (Face-A, Face-B and Face-C) for classroom/conference setting, which take as input $320\times240\times1$ monochromatic QVGA images. The Face-C model is a direct replication of the \rnnpool-Face-M4 model introduced in~\cite{rnnpool}. The architectures of these models are summarized in \textbf{\autoref{tab:faceDetectionArchitecture}} in \textbf{\autoref{sec:appendix}}. These \rnnpool-based models are designed for ARM Cortex-M class devices~\cite{rnnpoolBlog}, which support a minimum of $256$ KBs of SRAM and 1 MB of Flash. We train these models on the WIDER FACE~\cite{yang2016wider} dataset and fine-tune them on SCUT-HEAD~\cite{scut} dataset. For model accuracy, we evaluate the generated codes on 20\% validation split (405 images) of SCUT-HEAD Part-B, and report the MAP scores (\textbf{\autoref{sec:mlcvprelims}}).

\subsection{Comparison With 8-Bit Baselines}
\label{sec:8bitAccuracy}
We aim to justify our first claim by comparing the accuracy of models with different 8-bit number representations against the accuracy of the \fullprecision gold standard. \textbf{\autoref{fig:AccuracyDrop}} (data in \textbf{\autoref{tab:uniform8bitAccuracy}} of \textbf{\autoref{sec:appendix}}) shows that 8-bit representations have poor accuracy. In the case of 8-bit posits (\quarterprecisionposit), we simply pick the \textit{es} which empirically leads to better accuracy on a dataset. 8-bit fixed-point (\quarterprecision) gives the worst accuracy, 8-bit zero-skew representation of \tflite (\zskew) performs better, and \quarterprecisionposit performs the best among the 8-bit representations. However, even the accuracy loss of \quarterprecisionposit is unacceptable for \fastgrnn and \rnnpool. By using high bitwidth 16-bit posits sparingly, the accuracy drop of \toolposit is low for all models.

\subsection{Comparison With 16-Bit Baselines}
We compare the accuracy and RAM consumption of \toolposit against the gold-standard \fullprecision baseline. Since 16-bit representations also offer comparable accuracy with certain models at about $2 \times$ less RAM consumption, we compare with 16-bit fixed-point (\halfprecision) and \bfloat as well. We generate our \toolposit results while ensuring that the accuracy obtained stays within 0.2\% of the corresponding \fullprecision result, for each dataset.
\textbf{\autoref{fig:RAMCompression}}
 summarizes our experiments, where we achieve significant reduction of $2.9\times$--$5.1\times$ in peak RAM consumption even compared to \halfprecision or \bfloat, while observing only negligible differences in accuracy compared to \fullprecision. 
 Data for \textbf{\autoref{fig:RAMCompression}} is in \textbf{\autoref{tab:floatFixedPositAccuracy}} of \textbf{\autoref{sec:appendix}}.

\begin{figure*}
\centering
\begin{minipage}{\columnwidth}
  \centering
    \includegraphics[scale=0.79]{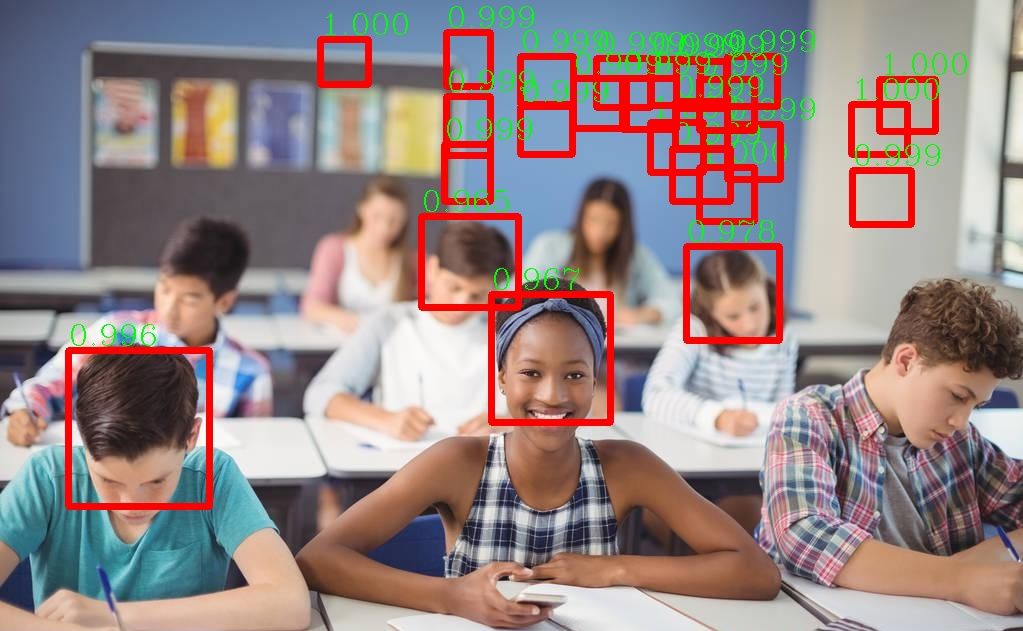}
    \caption{Example using Face-C model on \previoustoolfixed.}
    \label{fig:faceDetectionShiftry}
\end{minipage}
\hfill
\begin{minipage}{\columnwidth}
  \centering
    \includegraphics[scale=0.79]{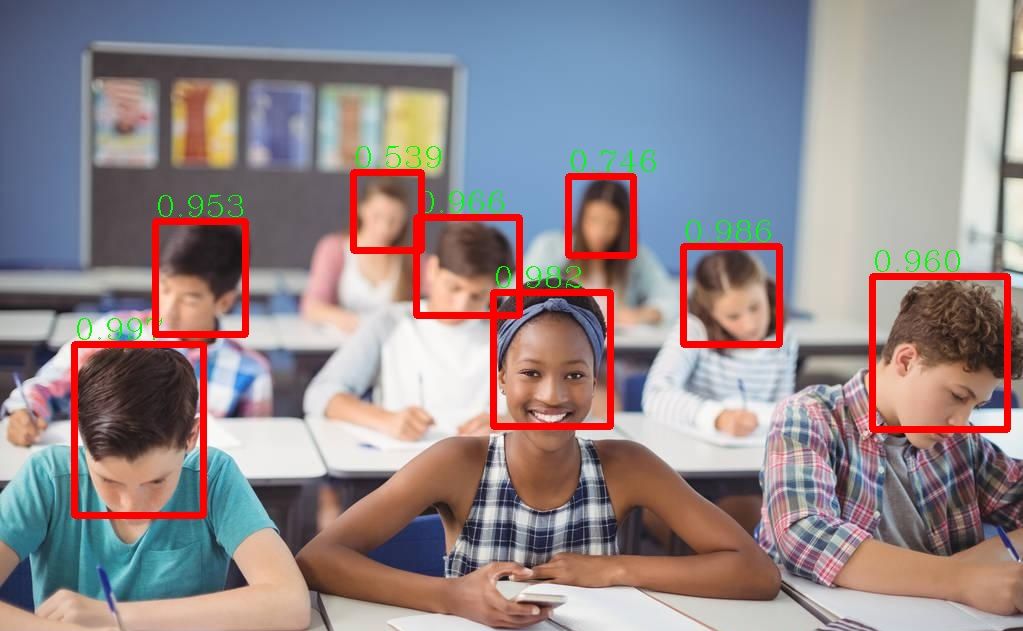}
    \caption{Example using Face-C model on \toolfixed.}
    \label{fig:faceDetectionMinUn}
\end{minipage}
\vspace{-10pt}
\end{figure*}

\subsection{Bitwidth Assignment}
We compare \tool's bitwidth assignment mechanism  with the mechanism of \previoustool and other exclusively accuracy-based and exclusively size-based mechanisms in both posit and fixed-point. 

In size-based baseline, we promote all tensors to higher bitwidth. Then, re-using \previoustool's cumulative demotion process, we demote individual tensors, by the order of their sizes, till we arrive at a bitwidth configuration which satisfies the memory constraint being considered. Since one can initiate cumulative demotion by either considering increasing or decreasing order of tensors sizes, we report the best out of those two possible results for each dataset.

Similarly, for accuracy-based baseline, we add the same memory constraint to \previoustool's demotion process. Here, we order the tensors by their individual demotion accuracies (i.e., the accuracy obtained by demoting a single tensor of interest to lower bitwidth, while keeping the remaining tensors in higher bitwidth) in decreasing order, and initiate the demotion process, and stop when we arrive at a bitwidth configuration which satisfies the limit. We also separately consider a tensor ``promotion'' experiment, where we initiate a cumulative promotion process after initializing all tensors to lower bitwidth, and promote tensors by the increasing order of their individual demotion accuracies, till we stay within the consumption limit.
Here as well, we report the best out of these two results for each dataset.

In all of our experiments, we consider the RAM consumption observed using \previoustool as the memory limit to be adhered to, and use soft limit factors from $\{1.0, 1.1\}$ for fixed-point, and $\{0.8, 0.9, 1.0, 1.1\}$ for our posit experiments. In case of \tool's result, we report the RAM consumption values while employing \previoustool's first-fit greedy heuristic, as well as the optimum memory management using Algorithm X. \textbf{\autoref{tab:fixedAccuracyBrief}} (full data in \textbf{\autoref{tab:fixedAccuracy}} of \textbf{\autoref{sec:appendix}})  summarizes our experiments for fixed-point representation, with average accuracy drops against \toolfixed mentioned for each exploration strategy on a per model basis. We observe better average accuracy numbers across most models and baselines while strictly staying within the same RAM constraints. \textbf{\autoref{sec:times}} shows that \tool achieves these improvements over \previoustool with negligible difference in execution latency, and a significant improvement in compilation latency. Similarly, \textbf{\autoref{tab:positAccuracyBrief}} (full data in \textbf{\autoref{tab:positAccuracy}} of \textbf{\autoref{sec:appendix}}) summarizes our experiments for posits, where we compare \previoustoolposit against \toolposit. Here as well, we observe better average accuracy numbers across all models and baselines while strictly staying within the same RAM constraints. Of particular interest is fixed-point \rnnpool  where \tool shows a 23.3\%
improvement in MAP score over \previoustool resulting in significant improvement in inference quality (see \textbf{\autoref{fig:faceDetectionShiftry}} and \textbf{\autoref{fig:faceDetectionMinUn}} for an example).


\input{tables/fixed_accuracy_ram_brief}

\input{tables/posit_accuracy_ram_brief}

\input{tables/fixed_point_times_brief}

\subsection{Inference Latency and Compilation Time}
\label{sec:times}
 Note that the RAM usage  and model accuracy are independent of the underlying hardware, and improvements in them (\textbf{\autoref{fig:AccuracyDrop}} and \textbf{\autoref{fig:RAMCompression}}) are oblivious to the microcontroller used. However, inference latency depends on the microcontroller and \tool generates code for a variety of microcontrollers.
\textbf{\autoref{tab:fixedPointTimesUno}} in \textbf{\autoref{sec:appendix}} provides the inference latency of running \tool-generated fixed-point classification models on Arduino Uno, a device with $2$ KB of RAM. Also see \textbf{\autoref{sec:microcontrollerprelims}} for a background on microcontrollers.
Further,  \textbf{\autoref{tab:fixedPointTimesBrief}} (full data in \textbf{\autoref{tab:fixedPointTimes}} of \textbf{\autoref{sec:appendix}}) compares inference latency of \tool and \previoustool. As expected, \tool's improvements in memory usage have negligible impact on the execution latency on microcontrollers.  For classification and localization models, the latency is measured on Arduino Due ($84$ MHz, $96$ KB SRAM, $512$ KB Flash) and STM32H747 ($240$ MHz, $1$ MB SRAM, $2$ MB Flash) boards respectively.

  We also compare the compilation time of \tool and \previoustool in \textbf{\autoref{tab:fixedPointTimesBrief}}.
 For a breakup of compilation times of \tool between bitwidth assignment and memory management, please refer to \textbf{\autoref{tab:fixedPointTimes}} of \textbf{\autoref{sec:appendix}}.
\subsection{Evaluation on ImageNet}
\label{sec:squeezenet}
\textbf{\autoref{tab:squeezenet}} evaluates \toolposit on SqueezeNet~\cite{squeezenet} - an ImageNet-scale CNN that matches the  accuracy of {\sc AlexNet} \cite{alexnet} while having $50\times$ smaller size. While SqueezeNet wasn't designed for being deployed on tiny devices, we demonstrate the versatility of our framework in achieving significant RAM usage reductions even for large models through this experiment. Note that for even larger CNNs like {\sc ResNet50} and {\sc MobileNets}, keeping all tensors in low bitwidth  maintains accuracy~\cite{tflite}, which makes bitwidth assignment trivial. To measure accuracy, we evaluate the generated code on $48,000$ RGB images spread across 1000 classes. We observe a peak RAM consumption of $1.44$ MBs on the code generated by the standard first-fit~\cite{nutt2002operating} heuristic for memory management on \toolposit, which was further reduced to $1.16$ MBs on using our optimum memory management, offering a improvement of almost $20\%$. As evident from \textbf{\autoref{tab:squeezenet}}, we achieve upto 3.81x reduction in peak RAM consumption with minimal loss in accuracy, relative to the \fullprecision gold standard, while existing TinyML frameworks (\tflite and \previoustool) fail to offer comparable reductions without leading to significant accuracy losses. In particular, \tool has $1.9\times$ reduction in RAM consumption over \previoustool while obtaining $6.5\%$ higher accuracy.

\subsection{Discussion}
Our evaluation shows that \tool significantly improves the inference quality and RAM usage over \previoustool without degrading inference latency or compilation time.
In particular, \textbf{\autoref{fig:faceDetectionShiftry}} and \textbf{\autoref{fig:faceDetectionMinUn}} show qualitatively the difference between the output of \tool and \previoustool on the image in \textbf{\autoref{fig:head}}.
These images show that improvements in MAP scores by \tool translates to significantly better inference output.
We discuss two approaches to improve \tool that we evaluated but did not pursue because the gains they provided were small in practice. 
\subsubsection{Permitting Overflows:}
For fixed-point, \haunter assigns bitwidths in a manner in which integer overflows are ruled out. However, this strategy is not always the best for achieving high classification accuracy. In particular, \previoustool permits values to overflow on some inputs to achieve higher precision on other inputs, so that the overall classification accuracy is high. While this optimization works out well in the case of fixed-point representation (where overflows are much better understood), it’s effect is hard to predict in case of other arbitrary number representations. Since \haunter doesn't permit overflows, it misses out on this trade-off. This limitation is best captured through \fastgrnn model on HAR-6 dataset (details in \textbf{\autoref{tab:fixedAccuracy}} of \textbf{\autoref{sec:appendix}}), where the generated model takes a hit of almost 2$\%$ compared to other strategies, and is the only cause of the average accuracy drops being negative for \fastgrnn models in \textbf{\autoref{tab:fixedAccuracyBrief}}. 

\input{tables/squeezenet_accuracy_ram}

\subsubsection{\textit{Generalization to More Bitwidth Options:}}
\label{sec:choice}
\haunter decides between two bitwidths, \textit{lowBitwidth} and \textit{highBitwidth}. It is plausible that if there are more bitwidth options then \haunter can generate even better code.
Having $k$ bitwidth options leads to a total of $k^N$ possible bitwidth assignments, where $N$ is the number of tensors in the program. In general, adding more bitwidth options directly into the algorithm blows up the search space and leads to many code executions, making \tool painfully slow. We experimented with three bitwidths ($8$, $12$ and $16$) for posits, but did not find any tangible benefit in terms of accuracy in our experiments compared to two bitwidth options of ($8$ and $16$) or ($8$ and $12$) or ($12$ and $16$) cases, depending on specific models.

As such, we chose a much simpler strategy in \textbf{\autoref{tab:floatFixedPositAccuracy}}. Given $B=\{8, 9, 10, 12, 16\}$ options for bitwidths, we generated $|B|$ homogeneous bitwidth codes (one for each option) and ordered the bitwidths by the accuracy/MAP obtained on their corresponding codes. Out of these sorted bitwidths, we simply chose the two options between which the 32-bit floating point accuracy lied, as the \textit{lowBitwidth} and \textit{highBitwidth} respectively. In situations where there was a tie in accuracy between multiple options for \textit{lowBitwidth} or \textit{highBitwidth}, we simply picked the lowest of the tied bitwidth options. In total, these only require $|B|$ additional code generations and executions, which is affordable since $|B| = 5$ is small.

%% file: figures/uniform_8_bit_accuracy.tex
\begin{figure}[H]
    \centering
    \graph
    {}
    {ymax=80}
    {ymin=-1}
    {\fastgrnn, \protonn, \bonsai, \rnnpool}
    {\addplot[pattern=horizontal lines, color=gray] coordinates {(\fastgrnn, 59.96) (\protonn, 6.34) (\bonsai, 5.41) (\rnnpool, 57.5)};
    \addplot[pattern=horizontal lines, color=yyellow] coordinates {(\fastgrnn, 11.30) (\protonn, 2.59) (\bonsai, -0.03) (\rnnpool, 23.9)};
    \addplot[pattern=crosshatch] coordinates {(\fastgrnn, 15.40) (\protonn, 0.22) (\bonsai, 0.37) (\rnnpool, 57.5)};
    \addplot[pattern=horizontal lines, color=black] coordinates {(\fastgrnn, -0.01) (\protonn, -0.06) (\bonsai, -0.02) (\rnnpool, 0.00)};}
    {\quarterprecision, \quarterprecisionposit, \zskew, \toolposit}
    {}
    {\fastgrnn}
    {\footnotesize Accuracy Drop $\%$}
    \caption{Accuracy/MAP drop against \fullprecision gold standard for different models (lower is better).}
    \label{fig:AccuracyDrop}
\end{figure}

%% file: figures/float_posit_ram.tex
\begin{figure}[H]
    \centering
    \graph
    {}
    {ymax=7}
    {ymin=0}
    {\fastgrnn, \protonn, \bonsai, \rnnpool}
    {\addplot[pattern=horizontal lines, color=gray] coordinates {(\fastgrnn, 1.0) (\protonn, 1.0) (\bonsai, 1.0) (\rnnpool, 1.0)};
    \addplot[pattern=horizontal lines, color=yyellow] coordinates {(\fastgrnn, 2.0) (\protonn, 2.0) (\bonsai, 2.0) (\rnnpool, 2.0)};
    \addplot[pattern=horizontal lines, color=black] coordinates {(\fastgrnn, 2.90) (\protonn, 4.13) (\bonsai, 5.13) (\rnnpool, 3.13)};}
    {\fullprecision, \halfprecision/\bfloat,  \toolposit}
    {}
    {\fastgrnn}
    {\footnotesize RAM Compression Factor}
    \caption{RAM compression relative to \fullprecision gold standard for different models (higher is better).}
    \label{fig:RAMCompression}
\end{figure}

%% file: tables/fixed_accuracy_ram_brief.tex
\begin{table}
\scriptsize
\caption{Average accuracy/MAP drop of computationally expensive exploration strategies and the previous state-of-the-art on \fastgrnn, \protonn, \bonsai and \rnnpool models against \toolfixed value.}
\vspace{3mm}
\centering
\begin{tabular}{c|c|c|c}
\multirow{2}{*}{\textbf{Model}} & \multicolumn{1}{c}{\textbf{\previoustoolfixed}} &  \multicolumn{1}{|c}{\textbf{Accuracy-Based}} & \multicolumn{1}{|c}{\textbf{Size-Based}}\\ 
 & \multicolumn{1}{c}{\textbf{Average Drop}} & \multicolumn{1}{|c}{\textbf{Average Drop}} & \multicolumn{1}{|c}{\textbf{Average Drop}}\\ 
\toprule
\multicolumn{1}{c|}{\textbf{\fastgrnn}} & \multicolumn{1}{c|}{0.5\%} & \multicolumn{1}{c|}{-0.2\%} & \multicolumn{1}{c}{-0.27\%}\\
\midrule
\multicolumn{1}{c|}{\textbf{\protonn}} & \multicolumn{1}{c|}{0.74\%} & \multicolumn{1}{c|}{0.48\%} & \multicolumn{1}{c}{0.71\%}\\
\midrule
\multicolumn{1}{c|}{\textbf{\bonsai}} & \multicolumn{1}{c|}{0.58\%} & \multicolumn{1}{c|}{0.30\%} & \multicolumn{1}{c}{0.25\%}\\
\midrule
\multicolumn{1}{c|}{\textbf{\rnnpool}} & \multicolumn{1}{c|}{23.3\%} & \multicolumn{1}{c|}{-0.3\%} & \multicolumn{1}{c}{-0.3\%}\\
\bottomrule
\end{tabular}
\vspace{-10pt}
\label{tab:fixedAccuracyBrief}
\end{table}

%% file: tables/posit_accuracy_ram_brief.tex
\begin{table}
\scriptsize
\caption{Average accuracy/MAP drop of computationally expensive exploration strategies and the previous state-of-the-art on \fastgrnn, \protonn, \bonsai and \rnnpool models against \toolposit value.}
\vspace{3mm}
\centering
\begin{tabular}{c|c|c|c}
\multirow{2}{*}{\textbf{Model}} & \multicolumn{1}{c}{\textbf{\previoustoolposit}} &  \multicolumn{1}{|c}{\textbf{Accuracy-Based}} & \multicolumn{1}{|c}{\textbf{Size-Based}}\\ 
 & \multicolumn{1}{c}{\textbf{Average Drop}} & \multicolumn{1}{|c}{\textbf{Average Drop}} & \multicolumn{1}{|c}{\textbf{Average Drop}}\\ 
\toprule
\multicolumn{1}{c|}{\textbf{\fastgrnn}} & \multicolumn{1}{c|}{0.58\%} & \multicolumn{1}{c|}{0.02\%} & \multicolumn{1}{c}{0.04\%}\\
\midrule
\multicolumn{1}{c|}{\textbf{\protonn}} & \multicolumn{1}{c|}{0.47\%} & \multicolumn{1}{c|}{0.17\%} & \multicolumn{1}{c}{1.73\%}\\
\midrule
\multicolumn{1}{c|}{\textbf{\bonsai}} & \multicolumn{1}{c|}{0.08\%} & \multicolumn{1}{c|}{0.14\%} & \multicolumn{1}{c}{0.06\%}\\
\midrule
\multicolumn{1}{c|}{\textbf{\rnnpool}} & \multicolumn{1}{c|}{3\%} & \multicolumn{1}{c|}{1.2\%} & \multicolumn{1}{c}{1.2\%}\\
\bottomrule
\end{tabular}
\vspace{-10pt}
\label{tab:positAccuracyBrief}
\end{table}

%% file: tables/fixed_point_times_brief.tex
\begin{table}
\scriptsize
\caption{Average inference latency and compilation time drop of \toolfixed  against \previoustoolfixed.}
\vspace{3mm}
\centering
\begin{tabular}{c|c|c}
\multirow{3}{*}{\textbf{Model}} & \multicolumn{1}{c}{\textbf{\toolfixed}} &  \multicolumn{1}{|c}{\textbf{\toolfixed}}\\
& \multicolumn{1}{c}{\textbf{Inference Latency}} & \multicolumn{1}{|c}{\textbf{Compilation Time}}\\
& \multicolumn{1}{c}{\textbf{Average  Drop}} & \multicolumn{1}{|c}{\textbf{Average  Drop}}\\
\toprule
\multicolumn{1}{c|}{\textbf{\fastgrnn}} & \multicolumn{1}{c|}{0.04\%} & \multicolumn{1}{c}{41.87\%}\\
\midrule
\multicolumn{1}{c|}{\textbf{\protonn}} & \multicolumn{1}{c|}{-7.02\%} & \multicolumn{1}{c}{33.25\%}\\
\midrule
\multicolumn{1}{c|}{\textbf{\bonsai}} & \multicolumn{1}{c|}{-10\%} & \multicolumn{1}{c}{44.57\%}\\
\midrule
\multicolumn{1}{c|}{\textbf{\rnnpool}} & \multicolumn{1}{c|}{0.03\%} & \multicolumn{1}{c}{-2.25\%}\\
\bottomrule
\end{tabular}
\vspace{-10pt}
\label{tab:fixedPointTimesBrief}
\end{table}

%% file: tables/squeezenet_accuracy_ram.tex
\begin{table}
\scriptsize
\caption{Evaluation of different frameworks on SqueezeNet.}
\vspace{3mm}
\centering
\begin{tabular}{c|c|c}
\textbf{Framework} & \textbf{Accuracy (\%)} & \textbf{RAM (MBs)}\\
\toprule
\textbf{\fullprecision} & 55.02 & 4.42\\
\midrule
\textbf{\previoustoolfixed (8-bit and 16-bit)} & 48.33 & 2.21\\
\midrule
\textbf{\tflite-\zskew} & 45.73 & 1.11\\
\midrule
\textbf{\toolposit (8-bit and 16-bit)} & 54.81 & 1.16\\
\bottomrule 
\end{tabular}
\vspace{-10pt}
\label{tab:squeezenet}
\end{table}

%% file: sections/9-conclusion.tex
\section{Conclusion}
\label{sec:concl}
\tool is the first TinyML framework which is parametric on number representation, generates a bitwidth assignment automatically taking into account both the size of the tensors and their impact on accuracy, and performs optimum memory management. Our evaluation shows that \tool significantly outperforms state-of-the-art TinyML frameworks in accuracy and RAM consumption.

%% file: sections/2-preliminaries.tex
\section{Preliminaries}
\label{sec:prelims}
In this section, we discuss the terminology from fixed-point representation, posit representation, machine learning and computer vision problems, and microcontrollers.

\subsection{Fixed-Point Representation}
\label{sec:fixedprelims}
In standard fixed-point arithmetic, a real number $r$ is stored as the $b$-bit signed integer $\lfloor r \times 2^{s}\rfloor_{b}$, where $s$ is a predetermined integer called the {\em scale} and $b$ is termed as the \textit{bitwidth} of the representation. As an example, consider the real number $r = 1.6181$:
\begin{flalign}
r = 1.6181 = 1.6181 \times \frac{2^{14}}{2^{14}} & \approx \frac{\lfloor{1.6181 \times 2^{14}}\rfloor}{2^{14}} \nonumber \\ & = 26510 \times 2^{-14} &&
\end{flalign}
To represent $1.6181$ in fixed-point arithmetic, we store the 16-bit integer $26510$, associated with the scale of 14. The interpreted real value would be $26510 \times 2^{-14} \approx 1.61804$, which is a close approximation of the original value.

For any real number and a given bitwidth, there is an optimum scale for the fixed-point representation; in the example above, given a 16-bit integer representation, 14 is the optimum scale for 1.6181, as using a scale of 15 or above causes integer overflow, and using a scale of 13 or below leads to more imprecise results.

A generalization of the above approach is the ``zero-skew'' quantization scheme by \tflite~\cite{tflite}. Here, a real number $r$ is stored in the following manner:
\begin{equation}
r = S(q_{b} - Z)
\end{equation}
where $S$ is an arbitrary positive real number, termed as the skew, and $Z$ is the quantized value of 0, termed as the zero-point, expressed in the same type as the $b$-bit unsigned integer $q_{b}$. Let us assume that we wish to represent a uniformly sampled real from the range $[-2, 2]$ in 8-bits. Hence, by choosing $r = 1.6181$, $S = \frac{\text{Variable Range}}{\text{uint8 Range}} = \frac{4}{255} \approx 0.015686$ and $Z = -\frac{\text{Variable Range Infimum}}{S} \approx 128$, we obtain $q$ as:
\begin{equation}
q_{8} = \lfloor{\frac{1.6181}{0.015686}}\rfloor + 128 \approx 231
\end{equation}

Substituting these approximations back into equation (1) gives us the interpreted value $1.6157$.

The skew and the zero-point are colloquially referred to as quantization parameters. All values in a tensor use a unique set of quantization parameters. A higher bitwidth preserves more precision, as long as the underlying integer does not overflow due to a high scale.

\subsection{Posit Representation}
\label{sec:positprelims}
Posit~\cite{gustafson1, gustafson2} is a novel real-number representation that is designed as an alternative to the IEEE-754 floating-point~\cite{ieee754fp}.

While an IEEE-754 float consists of a sign bit and a fixed number of bits for mantissa and exponent, the posit representation consists of 4 sections - the sign-bit ($s$), the regime bits ($r$), the exponent bits ($e$) and the fraction bits ($f$) - where the last three sections have a dynamic number of bits assigned to them, out of the overall fixed number of bits per posit.

\input{figures/posit_format}

The sign bit denotes the sign of the number, where a $0$ denotes a positive number and a $1$ denotes a negative number. The regime bits, which signify a super-exponent, define a regime value $k$, which can be positive, negative or zero:
\begin{itemize}
    \item Positive regime values are represented by a stream of $1$s delimited by a single $0$. The value of the regime would be one less than the number of leading $1$s. For example, if the regime bits are $1110$, then $k = 2$; $11111$ in a posit with only $6$ bits would correspond to $k = 4$.
    \item Negative regimes are represented by a stream of $0$s delimited by a single $1$. The value of the regime in this would be $k = -1 \times p$ where $p$ is the number of leading $0$s. For example, if the regime bits are $00001$, then $k = -4$; $000000$ in a posit with only $7$ bits would correspond to $k = -6$.
    \item The value zero is represented by the regime being $10$.
\end{itemize}

The length of the regime is dynamic, and is determined by the number of bits observed till a delimiting bit is encountered. Hence for an $n$-bit posit, there can be a minimum of $2$ and a maximum of $n - 1$ regime bits, (all bits excluding the sign bit). Therefore, the value of $k$ can range from $-(n - 1)$ to $(n - 2)$ in an $n$-bit number. For a given bitwidth, the \textit{es} value signifies the maximum number of bits available for the exponent section. Once the sign, regime and exponent bits have been considered, only then the remaining bits are interpreted as fraction bits.

To find the value of a number in posit representation, we define a parameter called \textit{useed} ($= 2^{2^{es}}$). The value of a posit with sign-bit $s$ is 
$$(-1)^s \times (useed)^k \times 2^E \times 1.F$$
where $E$ is the value of the exponent represented by the exponent bits ($e$), $k$ is the value denoted by the regime bits $r$, and $F$ is the fraction represented by the fraction bits $f$. It is possible in the posit representation for a number to contain no exponent bits and/or no fraction bits. In those cases the corresponding values of exponent ($E$) and fraction ($F$) are $0$.

For example, assuming $\textit{es} = 2$ and a bitwidth of 8 bits, a bit string $01101101$ will be interpreted as:
\begin{itemize}
    \item Sign bit ($s = 0$).
    \item Regime bits ($r = \{110\}$). Hence $k = 1$.
    \item Exponent bits ($e = \{11\}$). Hence $E = 2^{3} = 8$.
    \item Fraction bits ($f = \{01\}$). Hence $1.F = 1.25$.
\end{itemize}
Since $useed = 2^{2^{es}} = 16$, thus the real number represented by our bit string is $1 \times 16^1 \times 8 \times 1.25 = 160$.

\subsection{ML and Computer Vision Preliminaries}
\label{sec:mlcvprelims}
In this paper, we focus on two broad paradigms of ML problem space: classification and localization (commonly known as detection).

Classification-based models aim to predict singular discrete labels. For example, one can design a classifier which takes as input an image, and predicts a label indicating whether a human is present in the picture or not. The performance of such a model can be objectively measured using classification accuracy, which is the percentage of times the label is correct.

Localization-based models aim to predict more continuous outputs which determine the location of an instance. For example, given the image of a human being, one might need to employ a detection model in order to obtain the bounding-box coordinates of the human being in the picture frame. However, this makes the performance of the model continuous in terms of output quality. We use \textit{Mean Average Precision (MAP)} scores in such a scenario. MAP measures the mean area under the precision-recall curves, and is a standard metric for localization problems like face detection.

ML implementations typically choose the floating-point representation. However, different number representations offer salient advantages, in terms of computation, memory footprint and performance. As such, the effectiveness of a model expressed in a specific number representation is judged by how well it performs compared to its floating-point counterpart.

\subsection{Microcontrollers}
\label{sec:microcontrollerprelims}
Microcontrollers, designed for embedded applications, are a class of small, low-power computers on a single integrated circuit (IC). They comprise one or more CPU cores, with some memory and I/O peripherals. Two different kinds of memory are commonly used with microcontrollers, a reprogrammable {\it read-only} memory for storing firmware called Flash (similar to ROM, EPROM or EEPROM for personal computers) and a read-write memory called RAM (employing SRAM technology). Microcontrollers allow the storage of read-only bits such as model weights and the compiled software binaries on the Flash, and use their RAM for storing the mutable temporary variables corresponding to the results of intermediate computations. On these devices, RAM is a much more scarce resource than the Flash memory, with the latter often being user-expandable. Additionally, low-end commodity microcontrollers don't even employ hardware caches.

%% file: figures/posit_format.tex
\begin{figure}
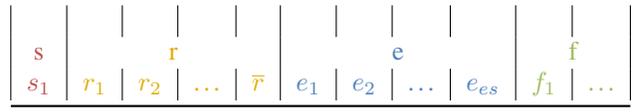

    \begin{tabular}{|c|c|c|c|c|c|c|c|c|c|c|}
    \toprule
    \multirow{3}{*}{\textcolor{rred}{s}}& & & & & & & & & &\\
    & \multicolumn{4}{c|}{\textcolor{yyellow}{r}} & \multicolumn{4}{c|}{\textcolor{bblue}{e}} & \multicolumn{2}{c|}{\textcolor{ggreen}{f}}\\
    \textcolor{rred}{$s_{1}$} & \textcolor{yyellow}{$r_{1}$} & \textcolor{yyellow}{$r_{2}$} & \textcolor{yyellow}{$\mydots$} & \textcolor{yyellow}{$\overline{r}$} & \textcolor{bblue}{$e_{1}$} & \textcolor{bblue}{$e_{2}$} & \textcolor{bblue}{$\mydots$} & \textcolor{bblue}{$e_{es}$} & \textcolor{ggreen}{$f_{1}$} & \textcolor{ggreen}{$\mydots$}\\
    \bottomrule
    \end{tabular}
    \caption{Example posit format.}
    \label{fig:positFormat}
\end{figure}

%% file: sections/3-working-example.tex
\section{Working Example}
\label{sec:workingExample}
To motivate the problem addressed in this paper, we provide an exemplar linear classifier model and generate the posit code for it. Consider \textbf{\autoref{code:float}}.
\input{codes/1-dsl-example}

Let us set a constraint for running this program in 3 bytes of peak RAM consumption. Here, we assume $W_1 \in \mathbb{R}^{1\times2}$, $X_1 \in \mathbb{R}^{2\times1}$ and $B_1 \in \mathbb{R}$ are read-only tensors. Hence, they are stored and fetched directly from the Flash, and do not count towards RAM usage. Additionally, we assume that the only two bitwidth options available to us are 8 bits and 16 bits. The corresponding 16-bit {\em homogeneous} (all tensors assigned the same bitwidth) posit code is depicted in \textbf{\autoref{code:posit}}. Here, only the temporary tensors $t_1$ and $t_2$ are stored in RAM.

\input{codes/2-half-word-posit-code}

We observe that the homogeneous 16-bit program uses 4 bytes of RAM (because of $t_1$ and $t_2$ being 2 bytes each), making it unsuitable for our target platform. While homogeneous 8-bit code would only take 2 bytes of RAM, and satisfy our memory constraint, it might lead to excessive losses in accuracy. Hence, we make use of {\em heterogeneous} bitwidth configurations for our tensors, which use 8-bit for some tensors and 16-bit for others, to reduce RAM consumption, while aiming to minimize the accuracy loss.

\haunter defines a promotability score for each tensor (\textbf{\autoref{tab:precSizePromotable}} and \textbf{\autoref{sec:heatMapGen}}). Ranking tensors in the decreasing order of promotability scores gives us an order list of promotion priority for the  different tensors in the program.

\input{tables/precision_sizes_promotability}
\input{tables/haunter_cumulative_promotion}
\input{codes/3-final-haunter-code}

Based on this promotion order, \haunter considers a few broad cases suitable for exploration, while starting from different initial configurations, as depicted in \textbf{\autoref{tab:cumulativePromotion}}, and explained in \textbf{\autoref{sec:promotionAlgo}}. At each step of an exploration instance, \haunter promotes a tensor cumulatively (i.e. in a successive operation on top of the previous bitwidth configuration), while strictly maintaining the constraint on the RAM consumption, and reverting the promotion in cases where the constraint is violated. At the end of each exploration instance, \haunter executes code on the derived bitwidth configuration, and records the accuracy.

Taking the example of the second instance from \textbf{\autoref{tab:cumulativePromotion}}, we initiate an exploration with all the tensors being assigned a low bitwidth. Then, based on the promotion order (i.e. $[t_2, t_1, X_1, B_1, W_1]$), we promote $t_2$ and generate the code on the new bitwidth configuration in order to estimate its RAM usage. We continue promoting other tensors in the promotion order, while keeping the previous tensors in the promoted state, till we end up promoting a tensor which leads to an overshoot of the memory limit ($t_1$). At this stage, we simply revert the promotion of the ``overshooting'' tensor, and continue to consider other tensors, till we have exhausted our promotion order. Once we have considered all tensors, we simply execute the generated code and record the accuracy.

Finally, \haunter chooses the configuration incurring the minimum loss in accuracy, and generates the final code based on this bitwidth configuration which keeps $t_2$ as an 8-bit posit and the rest remain as 16-bit posits. During memory management, the RAM tensors are mapped to indices into a global array called {\em scratch} (\textbf{\autoref{code:vbw}}) that resides in the RAM. This reduces the runtime overhead of allocating/deallocating tensors during program execution to zero. The RAM usage of the program at runtime is  upper-bounded by the size of {\em scratch}. Here, $t_1$ occupies the first two bytes, $t_2$ occupies the third byte of {\em scratch}, and the RAM consumption of \textbf{\autoref{code:vbw}} remains below 3-bytes, as desired.

%% file: codes/1-dsl-example.tex
\begin{procedure}
\caption{Linear Classifier()}
\label{code:float}
\DontPrintSemicolon
\small
    $W_1 := \begin{pmatrix}
        -2.139562 & 1.885351
        \end{pmatrix}$\;\vspace{1mm}
    $X_1 := \begin{pmatrix}
        1.185109 \\
        -2.206466
        \end{pmatrix}$\;\vspace{1mm}
    $B_1 := \begin{pmatrix}
        0.146048
        \end{pmatrix}$\;\vspace{1mm}
    $\mathbf{return}$ $(W_1 \times X_1) + B_1 $
\end{procedure}

%% file: codes/2-half-word-posit-code.tex
\begin{procedure}
\caption{16-bit Homogeneous() Posit (\textit{es}=2) Code}
\label{code:posit}
\DontPrintSemicolon
\small
    $\positg{16}{[1][2]} W_1 := \begin{pmatrix}
        -2.13965_{16} & 1.88525_{16}
        \end{pmatrix}$\;
    $\positg{16}{[2][1]} X_1 := \begin{pmatrix}
        1.18506_{16} \\
        -2.20605_{16}
        \end{pmatrix}$ \;
    $\positg{16}{[1][1]} B_1 := \begin{pmatrix}
        0.146057_{16}
        \end{pmatrix}$\;\;

    $\positg{16}{ }\  t_1,  t_2;$\;\;
    
    $t_1 = (W_1 \times X_1)$\;
    $t_2 = t_1 + B_1$\;
    
    $\mathbf{return}$ $t_2$
\end{procedure}

%% file: tables/precision_sizes_promotability.tex
\begin{table}[H]
\footnotesize
\caption{Variable sizes, values and promotability scores.}
\vspace{3mm}
\begin{tabular}{llll}
\toprule
\textbf{Var} & \textbf{16-Bit Values} & \textbf{8-Bit Values} & \textbf{Promotability}\\
\midrule
\vspace{1mm}
$W_1$ & (-2.13965, 1.88525)  & (-2.25, 1.875) & $0.00513$\\\vspace{1mm}
$B_1$ & 0.146057 & 0.140625 & $0.00543$\\\vspace{1mm}
$X_1$ & (1.18506, -2.20605) & (1.125, -2.25) & $0.02198$\\\vspace{1mm}
$t_1$ & -6.69531 & -7 & $0.30469$\\\vspace{1mm}
$t_2$ & -6.54883 & -7 & $0.45117$\\
\bottomrule
\end{tabular}
\label{tab:precSizePromotable}
\end{table}

%% file: tables/haunter_cumulative_promotion.tex
\begin{table}[H]
\footnotesize
\caption{Cumulative promotion within memory limit with different initial states chosen by \haunter. Repeated instances have been omitted for brevity.}
\vspace{3mm}
\centering
\begin{tabular}{lcc}
\toprule
\textbf{Promoted Vars} & \textbf{RAM Usage} & \textbf{Error} \\ 
\midrule
$None$ & 2 & 0.45047\\
\midrule
$t_2$ & 3 &\\
$t_2$, $t_1$ & 4 &\\
$t_2$, $X_1$ & 3 &\\
$t_2$, $X_1$, $B_1$ & 3 &\\
$t_2$, $X_1$, $B_1$, $W_1$ & 3 & 0.19601\\
\midrule
$t_1$ & 3 &\\
$t_1$, $t_2$ & 4 &\\
$t_1$, $X_1$ & 3 &\\
$t_1$, $X_1$, $B_1$ & 3 &\\
$t_1$, $X_1$, $B_1$, $W_1$ & 3 & 0.04953\\
\midrule
$t_2$, $t_1$, $X_1$, $B_1$, $W_1$ & 4 &\\
$t_1$, $X_1$, $B_1$, $W_1$ & 3 & 0.04953\\
\midrule
\vdots & &\\
\bottomrule
\end{tabular}
\label{tab:cumulativePromotion}
\end{table}

%% file: codes/3-final-haunter-code.tex
\begin{procedure}
\caption{Heterogeneous() Posit (\textit{es}=2) Code; $t_2$ uses 8-bits, rest use 16-bits}
\label{code:vbw}
\DontPrintSemicolon
\small
    $\positg{16}{[1][2]} W_1 := \begin{pmatrix}
        -2.13965_{16} & 1.88525_{16}
        \end{pmatrix}$\;
    $\positg{16}{[2][1]} X_1 := \begin{pmatrix}
        1.18506_{16} \\
        -2.20605_{16}
        \end{pmatrix}$ \;
    $\positg{16}{[1][1]} B_1 := \begin{pmatrix}
        0.146057_{16}
        \end{pmatrix}$\;\;
        

    
    
    $\mathtt{char}_{8}[3]\ \mathit{scratch};$\;\;
    
    $\mathit{scratch}[0:1] = \positg{16}{ }\left(W_1 \times X_1\right)$\;
    $\mathit{scratch}[2] = \positg{8}{ }\left(\mathit{scratch}[0:1] + B_1\right)$\;
    
    $\mathbf{return}$ $\mathit{scratch}[2]$
\end{procedure}

%% file: sections/12-haunter-extra-details.tex
\section{\haunter: Additional Details}
\label{sec:haunter-details}

\subsection{Complexity Analysis: Additional Details}
\label{sec:complexity-details}

We denote the amount of time required per code generation call as $\mathcal{T_{\mathit{codegen}}}$, and treat it as the upper bound of the code generation and compilation time required for different bitwidth configurations. We follow the same terminology for execution call latency (per example) as well, terming it $\mathcal{T_{\mathit{execution}}}$.

The bitwidth assignment mechanism of \previoustool follows an exploration phase where code generation, compilation and execution occurs twice per each tensor, once during individual tensor demotion, and once during cumulative tensor demotion. Hence for a model with $N$ tensors and supplied with a dataset of $D$ samples, \previoustool incurs a cost of $\mathcal{O}(N)$ code generation and $\mathcal{O}(N)$ execution calls, resulting in a bitwidth exploration latency of:
$$\mathcal{O}(N \times \mathcal{T_{\mathit{codegen}}} + N \times D \times \mathcal{T_{\mathit{execution}}})$$
\input{algorithms/1-haunter-preprocessing}
\input{algorithms/2-haunter-value-map}
For large models, supplied with large datasets, the second term is dominant, since $\mathcal{T_{\mathit{execution}}}$ becomes more expensive than $\mathcal{T_{\mathit{codegen}}}$. \haunter focuses on reducing the number of execution calls on average to $\mathcal{O}(\log{}N)$, thereby significantly reducing exploration latency.

The core observation that guides our average case analysis is that most ML models portray an asymmetric diversity in the tensor sizes. While classification or localization-based models try to ``downsize'' the input into smaller activations (by some multiplicative factor) at each step, generative models aim to ``upsize'' the input into larger activations. Empirically, for our benchmarks, the tensor sizes appear to follow an exponential distribution. With that assumption, a constant parameter $\alpha > 1$, and $\mathit{memoryLimit} = C$, for some positive scalar $C = \Theta(N^{\alpha}) << \alpha^{N}$,  we observe that the maximum number of tensors being, safely and cumulatively, promotable to a higher bitwidth is: $$\lfloor\log{(\mathit{memoryLimit}})\rfloor = \lfloor\log{C}\rfloor = \mathcal{O}(\log{N})$$

\input{algorithms/3-haunter-heat-map}
\input{algorithms/4-haunter-promotion-algorithm}

Hence, irrespective of the promotion order and the size of the $\mathit{overshootingVars}$ set in \textbf{\autoref{alg:attackingAlgo}}, only $\mathcal{O}(\log{}N)$ number of variables are able to proceed beyond line 10, leading to an average case of $\mathcal{O}(N\log{}N)$ code generation calls and $\mathcal{O}(\log{}N)$ execution calls in total. The total bitwidth exploration latency becomes:
$$\mathcal{O}(N \times \log{}N \times \mathcal{T_{\mathit{codegen}}} + \log{}N \times D \times \mathcal{T_{\mathit{execution}}})$$

\input{algorithms/5-haunter-attacking-algorithm}

This reduction in execution calls becomes significant even for moderately-sized models such as the Face-B model, described in \textbf{\autoref{sec:faceDetect}}, having 135 tensors, running on a dataset of 405 QVGA images. In our experiments, we observed the exploration time get cut from 49 minutes using \previoustoolfixed to 32 minutes using \toolfixed (\textbf{\autoref{tab:fixedPointTimes}} in \textbf{\autoref{sec:appendix}}).

%% file: algorithms/1-haunter-preprocessing.tex
\begin{algorithm}
\caption{Determining \textit{es} through accuracy.}
\label{alg:preprocessing}
\DontPrintSemicolon
\small
\SetKwFunction{FCompile}{Compile\&Execute}
\SetKwFunction{FPreprocess}{Preprocess}
\SetKwProg{Fn}{Function}{:}{}
    \Fn{\FPreprocess{$\mathit{lowBitwidth}, \mathit{highBitwidth}$}}{
    \nl    \If{$\mathit{lowBitwidth} ==  \mathit{8}$}{
    \nl        $\mathit{es8 = 2}$\;
    \nl        $\rho \leftarrow \{\mathit{var} \mapsto \mathit{8}\} \forall \mathit{var} \in \mathit{allVars}$\;
          \# Generate, compile and execute model code with the given set of parameters.\;
    \nl        $\mathit{accuracyES0} \leftarrow $ \FCompile{$\mathit{es=0}, \rho$}\;
    \nl        $\mathit{accuracyES2} \leftarrow $ \FCompile{$\mathit{es=2}, \rho$}\;
    \nl        \If{$\mathit{accuracyES0} >  \mathit{accuracyES2}$}{
    \nl            $\mathit{es8} = 0$}}
    \nl    \If{$\mathit{highBitwidth} ==  \mathit{16}$}{
    \nl        $\mathit{es16 = 2}$\;
    \nl        $\rho \leftarrow \{\mathit{var} \mapsto \mathit{16}\} \forall \mathit{var} \in \mathit{allVars}$\;
    \nl        $\mathit{accuracyES1} \leftarrow $ \FCompile{$\mathit{es=1}, \rho$}\;
    \nl        $\mathit{accuracyES2} \leftarrow $ \FCompile{$\mathit{es=2}, \rho$}\;
    \nl        \If{$\mathit{accuracyES1} >  \mathit{accuracyES2}$}{
    \nl            $\mathit{es16} = 1$}}
    \nl    \KwRet $\mathit{es8}, \mathit{es16}$\;
    }
\end{algorithm}

%% file: algorithms/2-haunter-value-map.tex
\begin{algorithm}
\caption{Generating value maps for each bitwidth.}
\label{alg:valMaps}
\DontPrintSemicolon
\small
\SetKwFunction{FCompile}{Compile\&Execute}
\SetKwFunction{FClearLog}{ClearLogs}
\SetKwFunction{FLog}{Log\&SaveValues}
\SetKwFunction{FSave}{SaveAcc\&Bitwidth}
\SetKwFunction{FCreateValueMaps}{CreateValueMaps}
\SetKwProg{Fn}{Function}{:}{}

    \Fn{\FCreateValueMaps{$\mathit{lowBitwidth}, \mathit{highBitwidth}$}}{
    \nl $\rho \leftarrow \{\mathit{var} \mapsto \mathit{highBitwidth}\} \forall \mathit{var} \in \mathit{allVars}$\;
    \# Clear logger file used for storing tensor values.\;
    \nl \FClearLog{}\;
    \nl $\mathit{accHigh} \leftarrow $ \FCompile{$\rho$}\;
    \# Create value map from logged data.\;
    \nl $\mathit{varsValueMapHigh} \leftarrow $ \FLog{}\;
    \nl $\rho \leftarrow \{\mathit{var} \mapsto \mathit{lowBitwidth}\} \forall \mathit{var} \in \mathit{allVars}$\;
    \nl \FClearLog{}\;
    \nl $\mathit{accLow} \leftarrow $ \FCompile{$\rho$}\;
    \nl $\mathit{varsValueMapLow} \leftarrow $ \FLog{}\;
    \# Save accuracy and bitwidth configuration information.\;
    \nl \FSave{$\mathit{accLow}, \rho$}\;
    \nl    \KwRet $\mathit{varsValueMapLow}, \mathit{varsValueMapHigh}$\;
    }
\end{algorithm}

%% file: algorithms/3-haunter-heat-map.tex
\begin{algorithm}
\caption{Generating heat map, given the value maps, and promotion order list.}
\label{alg:heatMap}
\DontPrintSemicolon
\small
\SetKwFunction{FPercentile}{Get95thPercentileValue}
\SetKwFunction{FCreateHeatMap}{CreateHeatMap}
\SetKwFunction{FArgSort}{DecreasingArgSort}
\SetKwProg{Fn}{Function}{:}{}

    \Fn{\FCreateHeatMap{$\mathit{varsValueMapLow}$, $\mathit{varsValueMapHigh}$}}{
    \nl    $\mathit{heatMap} = \{\}$\;
    \nl    \For{$\mathit{var} \in \mathit{allVars}$}{
    \nl        $\mathit{lowValues} \leftarrow \mathit{varsValueMapLow} [\mathit{var}]$\;
    \nl        $\mathit{highValues} \leftarrow \mathit{varsValueMapHigh} [\mathit{var}]$\;
    \nl        $\mathit{errors} \leftarrow []$\;
    \nl        \For{$\mathit{lowValue}, \mathit{highValue} \in \{\mathit{lowValues}, \mathit{highValues}\}$}{
    \nl            $\mathit{absError} = |\mathit{highValue} - \mathit{lowValue}|$\;
    \nl            $\mathit{error.append} (\mathit{absError})$\;}
    \nl        $\mathit{heatMap} [\mathit{var}] \leftarrow$  \FPercentile{$\mathit{errors}$} $\div \mathit{varsToSizeMap} [\mathit{var}]$\;
    }
    \nl $\mathit{promotionOrder} \leftarrow$ \FArgSort{$\mathit{heatMap}$}$\mathit{.keys()}$\;
    \nl    \KwRet $\mathit{heatMap}, \mathit{promotionOrder}$\;
    }
\end{algorithm}

%% file: algorithms/4-haunter-promotion-algorithm.tex
\begin{algorithm}
\caption{Helper function of \textbf{\autoref{alg:attackingAlgo}}; promoting all variables to high bitwidth subject to memory limit and soft limit factor.}
\label{alg:promotionAlgo}
\DontPrintSemicolon
\small
\SetKwFunction{FPromote}{PromoteWithinMemoryLimit}
\SetKwFunction{FPromotable}{CheckIfPromoted}
\SetKwFunction{FCompile}{Compile}
\SetKwProg{Fn}{Function}{:}{}

    \Fn{\FPromote{$\mathit{promotionOrder}$, $\mathit{memoryLimit}$, $\mathit{softLimitFactor}$}}{
    \nl    $\mathit{overshootingVars} \leftarrow []$\;
    \nl    \For{$\mathit{var} \in \mathit{promotionOrder}$}{
    \nl $\rho \leftarrow \{\mathit{var} \mapsto \mathit{highBitwidth}\}$\;
    \# Generate and compile model code for the given configuration.\;
    \nl        $\mathit{memoryUsage} \leftarrow $ \FCompile{$\rho$}\;
    \nl        \If{$\mathit{memoryUsage} > \mathit{memoryLimit} \times \mathit{softLimitFactor}$}{
    \nl            $\rho \leftarrow \{\mathit{var} \mapsto \mathit{lowBitwidth}\}$\;}
    \nl        \If{$\mathit{memoryUsage} \geq \mathit{memoryLimit} \times \mathit{softLimitFactor}$}{
    \nl            $\mathit{overshootingVars.append} (\mathit{var})$\;}
    }
    \nl    \KwRet $\mathit{overshootingVars}$\;
    }
\end{algorithm}

%% file: algorithms/5-haunter-attacking-algorithm.tex
\begin{algorithm}
\caption{\haunter Promotion Algorithm}
\label{alg:attackingAlgo}
\DontPrintSemicolon
\small
\SetKwFunction{FPromote}{PromoteWithinMemoryLimit}
\SetKwFunction{FAttacking}{PromotionAlgorithm}
\SetKwFunction{FGetMemoryUsage}{GetMemoryUsage}
\SetKwFunction{FCompileExec}{Compile\&Execute}
\SetKwFunction{FSave}{SaveAcc\&Bitwidth}
\SetKwFunction{FBest}{FindBitwidthConfigWithBestAccuracy}
\SetKwProg{Fn}{Function}{:}{}

    \Fn{\FAttacking{$\mathit{promotionOrder}$, $\mathit{memoryLimit}$, $\mathit{softLimitFactor}$}}{
    \nl    $\rho \leftarrow \{\mathit{var} \mapsto \mathit{lowBitwidth}\} \forall \mathit{var} \in \mathit{allVars}$\;
    \nl    $\mathit{overshootingVars} \leftarrow $ \FPromote{$\mathit{promotionOrder}$, $\mathit{memoryLimit}$, $\mathit{softLimitFactor}$}\;
    \nl    $\mathit{accuracy} \leftarrow $ \FCompileExec{$\rho$}\;
    \nl    \FSave{$\mathit{accuracy}, \rho$}\;
    \nl \;
    \nl    \For{$\mathit{promotable} \in \mathit{overshootingVars}$}{
    \nl        $\rho \leftarrow \{\mathit{var} \mapsto \mathit{lowBitwidth} \} \forall \mathit{var} \in \mathit{allVars}$\;
    \nl $\rho \leftarrow \{\mathit{promotable} \mapsto \mathit{highBitwidth}\}$\;
    \nl        $\mathit{memoryUsage} \leftarrow $ \FGetMemoryUsage{$\rho$}\;
    \nl        \If{$\mathit{memoryUsage} > \mathit{memoryLimit} \times \mathit{softLimitFactor}$}{
    \nl            $\mathit{continue}$\;}
    \nl        \FPromote{$\mathit{promotionOrder}$, $\mathit{memoryLimit}$, $\mathit{softLimitFactor}$}\;
    \nl        $\mathit{accuracy} \leftarrow $ \FCompileExec{$\rho$}\;
    \nl        \FSave{$\mathit{accuracy}, \rho$}
    }
    \nl \;
    \nl    $\rho \leftarrow \{\mathit{var} \mapsto \mathit{lowBitwidth} \} \forall \mathit{var} \in \mathit{allVars}$\;
    \nl    $\rho \leftarrow \{\mathit{var} \mapsto \mathit{highBitwidth} \} \forall \mathit{var} \in \mathit{overshootingVars}$\;
    \nl    \FPromote{$\mathit{promotionOrder}$, $\mathit{memoryLimit}$, $\mathit{softLimitFactor}$}\;
    \nl    $\mathit{accuracy} \leftarrow $ \FCompileExec{$\rho$}\;
    \nl    \FSave{$\mathit{accuracy}, \rho$}\;
    \nl    $\rho \leftarrow$ \FBest{}\;
    \nl    \KwRet $\rho$\;
    }
\end{algorithm}

%% file: sections/11-appendix.tex
\section{Evaluation Details}
\label{sec:appendix}
We now provide detailed results from all of our experiments.

\textbf{\autoref{tab:faceDetectionArchitecture}} summarizes the architecture of the face detection models evaluated in our experiments. All of our face detection models comprise more than one inverted bottleneck residual blocks, commonly called {\em MBConv} (see Fig. 3b of \cite{mobilenets}). Each MBConv layer consists of three convolution operations executed sequentially:

\begin{itemize}
    \item C1: A point-wise convolution which expands the number of channels from $\mathit{c_{in}}$ to $\mathit{c_{in}\times t}$, where $\mathit{t}$ is the expansion factor.
    \item C2: A depth-wise separable convolution with kernel size of $3\times3$ and stride of either $1$ or $2$.
    \item C3: A point-wise convolution which reduces the number of channels in the intermediate tensor from $\mathit{c_{in}\times t}$ to $\mathit{c_{out}}$.
\end{itemize}

\textbf{\autoref{tab:uniform8bitAccuracy}} provides accuracy values on different datasets, while comparing different 8-bit representations against the \fullprecision gold standard. For certain cases, the data-driven scale assignment mechanism of \previoustool is unable to determine the appropriate scales for 8-bit fixed-point numbers, and such instances are removed while calculating average accuracy drops.
These results show that 8-bit representations suffer from large losses in accuracy.

\textbf{\autoref{tab:floatFixedPositAccuracy}} provides accuracy and RAM consumption values on different datasets, while comparing different 16-bit representations, and the \fullprecision gold standard against \toolposit. The table shows that \toolposit suffers from  negligible differences in accuracy compared to \fullprecision, while achieving significant reduction in peak RAM consumption, even compared to \halfprecision or \bfloat.

\textbf{\autoref{tab:fixedAccuracy}} and \textbf{\autoref{tab:positAccuracy}} provide accuracy and RAM consumption values of different bitwidth assignment mechanisms for fixed-point and posit representation respectively. In both of these experiments, we set the RAM consumption of \previoustool's result as the memory limit for the other strategies. In case of \tool's result, we provide the RAM consumption values while employing the first-fit greedy heuristic, as well as the optimum memory management using Algorithm X.
This evaluation shows that \tool has better accuracy while consuming less memory than the baselines. Of particular interest is the FACE-B model in \textbf{\autoref{tab:fixedAccuracy}} where optimum RAM management reduces the RAM usage from $218$ KBs to $165$ KBs.

\textbf{\autoref{tab:fixedPointTimes}} (summarized by \textbf{\autoref{tab:fixedPointTimesBrief}}) provides the inference latency and compilation time of different models, on codes generated by \previoustoolfixed and \toolfixed corresponding to \textbf{\autoref{tab:fixedAccuracy}} (summarized by \textbf{\autoref{tab:fixedAccuracyBrief}}). Inference latency for classification and localization models has been generated on Arduino Due ($84$ MHz, $96$ KB SRAM, $512$ KB Flash) and STM32H747 ($240$ MHz, $1$ MB SRAM, $2$ MB Flash) boards respectively. This evaluation shows that the reduction in RAM usage by \tool does not degrade latency on microcontrollers. \textbf{\autoref{tab:fixedPointTimesUno}} provides the inference latency obtained on Arduino Uno ($16$ MHz, $2$ KB SRAM, $32$ KB Flash) for \toolfixed experiments.

\raggedbottom

\input{tables/face_detection_architecture}

\input{tables/uniform_8_bit_accuracy}

\input{tables/float_fixed_posit_accuracy_ram}

\input{tables/fixed_accuracy_ram}

\input{tables/posit_accuracy_ram}

\input{tables/fixed_point_times}

\input{tables/fixed_point_times_uno}

%% file: tables/face_detection_architecture.tex
\begin{table}[H]
\footnotesize
\caption{Architecture schematics of face detection models. Each line denotes a sequence of layers, repeated $n$ times. The first layer of each MBConv sequence has stride $s$ and rest use stride $1$. Expansion factor $t$ is multiplied to the input channels to change the width. The number of output channels is $c$.}
\vspace{3mm}
\centering
\begin{tabular}{c|c|c|c|c|c}
\textbf{Input} & \textbf{Layer} & \textbf{t} &  \textbf{c} & \textbf{n} & \textbf{s}\\
\toprule
$320 \times 240 \times 1$ & Conv2D $3 \times 3$ & 1 & 4 & 1 & 2\\
$160 \times 120 \times 4$ & RNNPool & 1 & 64 & 1 & 4\\
$40 \times 30 \times 64$ & MBConv & 2 & 32 & 1 & 1\\
$40 \times 30 \times 32$ & MBConv & 2 & 32 & 7 & 1\\
$40 \times 30 \times 32$ & MBConv & 2 & 96 & 1 & 2\\
$20 \times 15 \times 96$ & MBConv & 2 & 96 & 2 & 1\\
$20 \times 15 \times 96$ & MBConv & 2 & 128 & 1 & 1\\
$20 \times 15 \times 128$ & MBConv & 2 & 128 & 2 & 1\\
\bottomrule 
\end{tabular}
\caption*{The architecture of Face-A.}
\vspace{3mm}
\begin{tabular}{c|c|c|c|c|c}
\textbf{Input} & \textbf{Layer} & \textbf{t} &  \textbf{c} & \textbf{n} & \textbf{s}\\
\toprule
$320 \times 240 \times 1$ & Conv2D $3 \times 3$ & 1 & 4 & 1 & 2\\
$160 \times 120 \times 4$ & RNNPool & 1 & 64 & 1 & 8\\
$20 \times 15 \times 64$ & MBConv & 2 & 32 & 1 & 1\\
$20 \times 15 \times 32$ & MBConv & 2 & 96 & 1 & 1\\
$20 \times 15 \times 96$ & MBConv & 2 & 96 & 1 & 1\\
$20 \times 15 \times 96$ & MBConv & 2 & 128 & 1 & 1\\
\bottomrule
\end{tabular}
\caption*{The architecture of Face-B.}
\vspace{3mm}
\begin{tabular}{c|c|c|c|c|c}
\textbf{Input} & \textbf{Layer} & \textbf{t} &  \textbf{c} & \textbf{n} & \textbf{s}\\
\toprule
$320 \times 240 \times 1$ & Conv2D $3 \times 3$ & 1 & 4 & 1 & 2\\
$160 \times 120 \times 4$ & RNNPool & 1 & 64 & 1 & 4\\
$40 \times 30 \times 64$ & MBConv & 2 & 32 & 1 & 1\\
$40 \times 30 \times 32$ & MBConv & 2 & 32 & 1 & 1\\
$40 \times 30 \times 32$ & MBConv & 2 & 64 & 1 & 2\\
$20 \times 15 \times 64$ & MBConv & 2 & 64 & 1 & 1\\
\bottomrule
\end{tabular}
\caption*{The architecture of Face-C.}
\label{tab:faceDetectionArchitecture}
\end{table}

%% file: tables/uniform_8_bit_accuracy.tex
\begin{table*}
\footnotesize
\caption{Performance of homogeneous 8-bit codes of different representations on \fastgrnn, \protonn, \bonsai and \rnnpool models against the \fullprecision gold standard. Accuracy is in percent. For classification-based models, the number of classes follow the dataset name. An $\times$ in the column indicates that data-driven compilation fails to determine a suitable scale. Average accuracy/MAP drop is calculated with respect to \fullprecision value.}
\vspace{3mm}
\centering
\begin{tabular}{c|c|c|c|c}
\multicolumn{1}{c}{\textbf{Model \&}} & \multicolumn{1}{|c}{\textbf{\fullprecision}} &
\multicolumn{1}{|c}{\textbf{\quarterprecision}} &
\multicolumn{1}{|c}{\textbf{\quarterprecisionposit (Best ES)}} &
\multicolumn{1}{|c}{\textbf{\tflite-\zskew}} \\ 
\multicolumn{1}{c}{\textbf{Dataset}} & \multicolumn{1}{|c}{\textbf{Accuracy}} & \multicolumn{1}{|c}{\textbf{Accuracy}} & \multicolumn{1}{|c}{\textbf{Accuracy}} & \multicolumn{1}{|c}{\textbf{Accuracy}}\\
\toprule
\multicolumn{1}{c|}{\textbf{\fastgrnn}} & & & &\\
DSA-19 & 77.76 & 6.56 & 69.87 & 76.56\\
WAKEWORD-2 & 98.97 & 95.66 & 98.52 & 84.13\\
GOOGLE-12 & 92.99 & 8.08 & 44.63 & 84.42\\
GOOGLE-30 & 78.74 & 4.21 & 51.57 & 58.92\\
HAR-2 & 91.65 & 50.87 & 84.83 & 88.12\\
HAR-6 & 92.03 & 16.36 & 91.31 & 91.89\\
USPS-10 & 94.12 & 52.12 & 93.92 & 93.87\\
MNIST-10 & 98.05 & $\times$ & 96.45 & 12.04\\
INDUSTRIAL-72 & 89.96 & 2.73 & 81.50 & 85.69\\
\midrule
\multicolumn{1}{c|}{\textbf{Average Drop}} & \multicolumn{1}{c|}{\textbf{Base Case}} & \multicolumn{1}{c|}{\textbf{59.96}} & \multicolumn{1}{c|}{\textbf{11.30}} & \multicolumn{1}{c}{\textbf{15.40}}\\
\midrule
\multicolumn{1}{c|}{\textbf{\protonn}} & & & &\\
CIFAR-2 & 76.45 & 71.47 & 66.18 & 76.64\\
CR-2 & 72.91 & 69.72 & 66.33 & 72.85\\
CR-62 & 32.74 & 30.76 & 31.33 & 31.75\\
CURET-61 & 54.53 & 39.42 & 53.10 & 54.17\\
LETTER-26 & 84.02 & 80.84 & 82.80 & 83.50\\
MNIST-2 & 95.21 & 87.40 & 94.11 & 95.23\\
MNIST-10 & 92.06 & 85.98 & 91.23 & 91.78\\
USPS-2 & 94.27 & 92.58 & 93.32 & 94.27\\
USPS-10 & 92.53 & 90.78 & 91.83 & 92.73\\
WARD-2 & 94.87 & 77.21 & 93.42 & 94.51\\
\midrule
\multicolumn{1}{c|}{\textbf{Average Drop}} & \multicolumn{1}{c|}{\textbf{Base Case}} & \multicolumn{1}{c|}{\textbf{6.34}} & \multicolumn{1}{c|}{\textbf{2.59}} & \multicolumn{1}{c}{\textbf{0.22}}\\
\midrule
\multicolumn{1}{c|}{\textbf{\bonsai}} & & & &\\
CIFAR-2 & 75.95 & 73.14 & 75.90 & 75.70\\
CR-2 & 74.60 & 72.22 & 74.97 & 74.34\\
CR-62 & 10.76 & 9.03 & 10.76 & 10.03\\
CURET-61 & 45.55 & 34.50 & 45.12 & 44.05\\
LETTER-26 & 65.04 & 60.96 & 65.28 & 64.62\\
MNIST-2 & 95.96 & 93.51 & 95.88 & 96.03\\
MNIST-10 & 93.56 & 76.32 & 93.49 & 93.28\\
USPS-2 & 94.82 & 91.08 & 94.92 & 94.67\\
USPS-10 & 91.63 & 89.89 & 91.83 & 91.68\\
WARD-2 & 95.13 & 88.30 & 95.18 & 94.93\\
\midrule
\multicolumn{1}{c|}{\textbf{Average Drop}} & \multicolumn{1}{c|}{\textbf{Base Case}} & \multicolumn{1}{c|}{\textbf{5.41}} & \multicolumn{1}{c|}{\textbf{-0.03}} & \multicolumn{1}{c}{\textbf{0.37}}\\
\midrule
\multicolumn{1}{c|}{\textbf{\rnnpool}} & & & &\\
FACE-A & 0.643 & x & 0.560 & x\\
FACE-B & 0.336 & x & 0.209 & x\\
FACE-C & 0.575 & 5e-5 & 0.069 & 3e-5\\
\midrule
\multicolumn{1}{c|}{\textbf{Average Drop}} & \multicolumn{1}{c|}{\textbf{Base Case}} & \multicolumn{1}{c|}{\textbf{0.575}} & \multicolumn{1}{c|}{\textbf{0.239}} & \multicolumn{1}{c}{\textbf{0.575}}\\
\bottomrule
\end{tabular}
\label{tab:uniform8bitAccuracy}
\end{table*}

%% file: tables/float_fixed_posit_accuracy_ram.tex
\begin{table*}
\footnotesize
\caption{Performance of \toolposit on \fastgrnn, \protonn, \bonsai and \rnnpool models against \fullprecision, \halfprecision and \bfloat baselines. Accuracy is in percent, and size is in bytes. For classification-based models, the number of classes follow the dataset name. Average accuracy/MAP drop is calculated with respect to the \fullprecision value.}
\vspace{3mm}
\centering
\begin{tabular}{c|cc|cc|cc|ccc}
\multicolumn{1}{c|}{\textbf{Model}} & \multicolumn{2}{|c}{\textbf{\fullprecision}} & \multicolumn{2}{|c}{\textbf{\halfprecision}} & \multicolumn{2}{|c}{\textbf{\bfloat}} &
\multicolumn{3}{|c}{\textbf{\toolposit}}\\ 
\multicolumn{1}{c|}{\textbf{\&}} & \multirow{2}{*}{\textbf{Accuracy}} & \multirow{2}{*}{\textbf{RAM}} & \multirow{2}{*}{\textbf{Accuracy}} & \multirow{2}{*}{\textbf{RAM}} & \multirow{2}{*}{\textbf{Accuracy}} & \multirow{2}{*}{\textbf{RAM}} &\multirow{2}{*}{\textbf{Accuracy}} & \multicolumn{1}{c}{\textbf{RAM}} & \multirow{2}{*}{\textbf{Bitwidth}}\\
\multicolumn{1}{c|}{\textbf{Dataset}} & & & & & & & & \multicolumn{1}{c}{\textbf{(OPT)}}\\
\toprule
\multicolumn{1}{c|}{\textbf{\fastgrnn}} & & & & & & & &\\
DSA-19 & 77.76 & 1280 & 77.70 & 640 & 77.59 & 640 & 78.00 & 448 & \{8, 12\}\\
WAKEWORD-2 & 98.97 & 640 & 98.74 & 320 & 98.97 & 320 & 99.20 & 192 & \{8, 12\}\\
GOOGLE-12 & 92.99 & 2000 & 92.99 & 1000 & 92.76 & 1000 & 92.83 & 700 & \{10, 12\}\\
GOOGLE-30 & 78.74 & 2000 & 78.86 & 1000 & 78.76 & 1000 & 78.70 & 750 & \{8, 12\}\\
HAR-2 & 91.65 & 1600 & 91.52 & 800 & 91.55 & 800 & 91.55 & 500 & \{8, 10\}\\
HAR-6 & 92.03 & 1600 & 92.03 & 800 & 91.52 & 800 & 91.99 & 560 & \{8, 12\}\\
USPS-10 & 94.12 & 1664 & 94.12 & 832 & 94.12 & 832 & 94.17 & 448 & \{8, 10\}\\
MNIST-10 & 98.05 & 2560 & 85.74 & 1280 & 98.12 & 1280 & 98.10 & 960 & \{8, 12\}\\
INDUSTRIAL-72 & 89.96 & 1280 & 89.88 & 640 & 89.53 & 640 & 89.80 & 480 & \{12, 16\}\\
\midrule
\multicolumn{1}{c|}{\textbf{Average Drop}} & \multicolumn{2}{c|}{\textbf{Base Case}} & \multicolumn{2}{c|}{\textbf{0.41}} & \multicolumn{2}{c|}{\textbf{0.15}} & \multicolumn{3}{c}{\textbf{-0.01}}\\
\midrule
\multicolumn{1}{c|}{\textbf{\protonn}} & & & & & & & &\\
CIFAR-2 & 76.45 & 312 & 76.49 & 156 & 76.47 & 156 & 76.59 & 70 & \{8, 12\}\\
CR-2 & 72.91 & 412 & 73.07 & 206 & 73.12 & 206 & 73.17 & 94 & \{8, 12\}\\
CR-62 & 32.74 & 552 & 32.74 & 276 & 32.79 & 276 & 32.69 & 171 & \{8, 12\}\\
CURET-61 & 54.53 & 544 & 54.53 & 272 & 54.67 & 272 & 54.53 & 138 & \{8, 12\}\\
LETTER-26 & 84.02 & 308 & 83.60 & 154 & 83.92 & 154 & 84.02 & 81 & \{8, 12\}\\
MNIST-2 & 95.21 & 240 & 95.23 & 120 & 95.25 & 120 & 95.30 & 39 & \{8, 10\}\\
MNIST-10 & 92.06 & 244 & 92.04 & 122 & 92.09 & 122 & 92.10 & 49 & \{8, 10\}\\
USPS-2 & 94.27 & 240 & 94.27 & 120 & 94.32 & 120 & 94.47 & 44 & \{8, 12\}\\
USPS-10 & 92.53 & 3240 & 92.58 & 1620 & 92.58 & 1620 & 92.48 & 800 & \{8, 9\}\\
WARD-2 & 94.87 & 212 & 94.87 & 106 & 94.82 & 106 & 94.82 & 39 & \{8, 10\}\\
\midrule
\multicolumn{1}{c|}{\textbf{Average Drop}} & \multicolumn{2}{c|}{\textbf{Base Case}} & \multicolumn{2}{c|}{\textbf{0.02}} & \multicolumn{2}{c|}{\textbf{-0.05}} & \multicolumn{3}{c}{\textbf{-0.06}}\\
\midrule
\multicolumn{1}{c|}{\textbf{\bonsai}} & & & & & & & &\\
CIFAR-2 & 75.95 & 320 & 75.94 & 160 & 75.91 & 160 & 75.90 & 40 & \{8, 12\}\\
CR-2 & 74.60 & 320 & 74.66 & 160 & 74.71 & 160 & 74.97 & 40 & \{8, 12\}\\
CR-62 & 10.76 & 1072 & 10.81 & 536 & 10.76 & 536 & 10.76 & 258 & \{8, 12\}\\
CURET-61 & 45.55 & 1072 & 45.55 & 536 & 45.62 & 536 & 45.55 & 255 & \{8, 12\}\\
LETTER-26 & 65.04 & 496 & 65.10 & 248 & 64.98 & 248 & 64.84 & 114 & \{8, 12\}\\
MNIST-2 & 95.96 & 160 & 95.96 & 80 & 95.96 & 80 & 95.98 & 20 & \{8, 12\}\\
MNIST-10 & 93.56 & 240 & 93.47 & 120 & 93.51 & 120 & 93.44 & 50 & \{8, 12\}\\
USPS-2 & 94.82 & 480 & 94.82 & 240 & 94.82 & 240 & 94.92 & 60 & \{8, 12\}\\
USPS-10 & 91.63 & 304 & 91.63 & 152 & 91.53 & 152 & 91.63 & 45 & \{8, 12\}\\
WARD-2 & 95.13 & 160 & 95.18 & 80 & 95.13 & 80 & 95.18 & 20 & \{8, 12\}\\
\midrule
\multicolumn{1}{c|}{\textbf{Average Drop}} & \multicolumn{2}{c|}{\textbf{Base Case}} & \multicolumn{2}{c|}{\textbf{-0.01}} & \multicolumn{2}{c|}{\textbf{0.01}} & \multicolumn{3}{c}{\textbf{-0.02}}\\
\midrule
\multicolumn{1}{c|}{\textbf{\rnnpool}} & & & & & & & &\\
FACE-A & 0.643 & 618K & 0.628 & 309K & 0.646 & 309K & 0.647 & 197K & \{10, 12\}\\
FACE-B & 0.336 & 614K & 0.301 & 307K & 0.335 & 307K & 0.333 & 192K & \{10, 12\}\\
FACE-C & 0.575 & 618K & 0.562 & 309K & 0.574 & 309K & 0.573 & 203K & \{10, 12\}\\
\midrule
\multicolumn{1}{c|}{\textbf{Average Drop}} & \multicolumn{2}{c|}{\textbf{Base Case}} & \multicolumn{2}{c|}{\textbf{0.021}} & \multicolumn{2}{c|}{\textbf{0.000}} & \multicolumn{3}{c}{\textbf{0.000}}\\
\bottomrule
\end{tabular}
\label{tab:floatFixedPositAccuracy}
\end{table*}

%% file: tables/fixed_accuracy_ram.tex
\begin{table*}
\footnotesize
\caption{Performance of \toolfixed on \fastgrnn, \protonn, \bonsai and \rnnpool models against computationally expensive bitwidth assignment mechanisms and the previous state-of-the-art. Accuracy is in percent, and size is in bytes. For classification-based models, the number of classes follow the dataset name. Average accuracy/MAP drop is calculated with respect to \toolfixed value.}
\vspace{3mm}
\centering
\begin{tabular}{c|cc|cc|cc|ccc}
\multicolumn{1}{c|}{\textbf{Model}} & \multicolumn{2}{c}{\textbf{\previoustoolfixed}} &  \multicolumn{2}{|c}{\textbf{Accuracy-Based}} & \multicolumn{2}{|c}{\textbf{Size-Based}} &
\multicolumn{3}{|c}{\textbf{\toolfixed}}\\ 
\multicolumn{1}{c|}{\textbf{\&}} & \multirow{2}{*}{\textbf{Accuracy}} & \multirow{2}{*}{\textbf{RAM}} & \multirow{2}{*}{\textbf{Accuracy}} & \multirow{2}{*}{\textbf{RAM}} & \multirow{2}{*}{\textbf{Accuracy}} & \multirow{2}{*}{\textbf{RAM}} & \multirow{2}{*}{\textbf{Accuracy}} & \multirow{2}{*}{\textbf{RAM}} & \multicolumn{1}{c}{\textbf{RAM}}\\
\multicolumn{1}{c|}{\textbf{Dataset}} & & & & & & & & & \multicolumn{1}{c}{\textbf{(OPT)}}\\
\toprule
\multicolumn{1}{c|}{\textbf{\fastgrnn}} & & & & & & & & & \\
DSA-19 & 74.39 & 740 & \textbf{77.70} & 640 & \textbf{77.70} & 640 & \textbf{77.70} & 640 & 640\\
WAKEWORD-2 & 98.63 & 377 & \textbf{98.74} & 320 & \textbf{98.74} & 320 & \textbf{98.74} & 320 & 320\\
GOOGLE-12 & 92.24 & 1141 & \textbf{92.99} & 1000 & \textbf{92.99} & 1000 & \textbf{92.99} & 1000 & 1000\\
GOOGLE-30 & 78.42 & 1000 & \textbf{78.86} & 1000 & \textbf{78.86} & 1000 & \textbf{78.86} & 1000 & 1000\\
HAR-2 & 90.09 & 845 & \textbf{91.52} & 800 & \textbf{91.52} & 800 & \textbf{91.52} & 800 & 800\\
HAR-6 & 90.23 & 720 & 90.50 & 720 & \textbf{90.84} & 720 & 88.26 & 730 & 720\\
USPS-10 & 92.98 & 512 & 92.98 & 512 & 92.88 & 512 & \textbf{93.02} & 512 & 512\\
MNIST-10 & \textbf{85.74} & 1280 & \textbf{85.74} & 1280 & \textbf{85.74} & 1280 & \textbf{85.74} & 1280 & 1280\\
INDUSTRIAL-72 & 87.31 & 520 & 87.31 & 520 & \textbf{87.71} & 520 & \textbf{87.71} & 520 & 520\\
\midrule
\multicolumn{1}{c|}{\textbf{Average Drop}} & \multicolumn{2}{c|}{\textbf{0.5}} & \multicolumn{2}{c|}{\textbf{-0.2}} & \multicolumn{2}{c|}{\textbf{-0.27}} & \multicolumn{3}{c}{\textbf{Base Case}}\\
\midrule
\multicolumn{1}{c|}{\textbf{\protonn}} & & & & & & & & & \\
CIFAR-2 & 75.18 & 93 & 75.65 & 93 & 75.43 & 81 & \textbf{76.16} & 93 & 93\\
CR-2 & \textbf{72.85} & 124 & 72.53 & 124 & 71.32 & 106 & 72.64 & 124 & 124\\
CR-62 & 31.70 & 272 & 31.70 & 272 & 31.75 & 226 & \textbf{32.27} & 226 & 226\\
CURET-61 & 53.10 & 146 & \textbf{54.10} & 145 & 51.60 & 138 & 53.89 & 145 & 145\\
LETTER-26 & 81.78 & 77 & 81.78 & 77 & 82.42 & 77 & \textbf{83.30} & 77 & 77\\
MNIST-2 & 94.80 & 60 & 95.07 & 60 & 95.12 & 60 & \textbf{95.26} & 60 & 56\\
MNIST-10 & 91.78 & 61 & \textbf{91.95} & 61 & 91.93 & 61 & 91.94 & 61 & 61\\
USPS-2 & 92.83 & 60 & 93.92 & 60 & 94.02 & 60 & \textbf{94.07} & 60 & 56\\
USPS-10 & 91.63 & 810 & 91.63 & 810 & 92.03 & 810 & \textbf{92.73} & 820 & 800\\
WARD-2 & 93.68 & 63 & 93.68 & 63 & 94.05 & 63 & \textbf{94.51} & 63 & 63\\
\midrule
\multicolumn{1}{c|}{\textbf{Average Drop}} & \multicolumn{2}{c|}{\textbf{0.74}} & \multicolumn{2}{c|}{\textbf{0.48}} & \multicolumn{2}{c|}{\textbf{0.71}} & \multicolumn{3}{c}{\textbf{Base Case}}\\
\midrule
\multicolumn{1}{c|}{\textbf{\bonsai}} & & & & & & & & & \\
CIFAR-2 & 74.60 & 80 & 74.60 & 80 & 74.89 & 80 & \textbf{75.08} & 80 & 66\\
CR-2 & 74.34 & 100 & 74.34 & 100 & \textbf{74.87} & 100 & 74.44 & 100 & 66\\
CR-62 & 9.40 & 268 & 9.40 & 268 & 9.40 & 268 & \textbf{9.56} & 288 & 268\\
CURET-61 & 42.77 & 524 & 43.41 & 524 & 43.98 & 524 & \textbf{45.55} & 536 & 488\\
LETTER-26 & 63.26 & 124 & 63.48 & 108 & 63.52 & 108 & \textbf{63.52} & 122 & 108\\
MNIST-2 & 95.85 & 50 & \textbf{96.03} & 50 & 95.99 & 50 & 95.92 & 50 & 36\\
MNIST-10 & 92.70 & 60 & \textbf{93.12} & 60 & 92.59 & 60 & 93.05 & 60 & 60\\
USPS-2 & 94.07 & 120 & 94.07 & 120 & 94.12 & 120 & \textbf{94.37} & 120 & 96\\
USPS-10 & 91.48 & 76 & 91.48 & 76 & 91.53 & 76 & \textbf{91.63} & 80 & 72\\
WARD-2 & 93.94 & 50 & \textbf{95.29} & 50 & 94.77 & 50 & 95.08 & 50 & 36\\
\midrule
\multicolumn{1}{c|}{\textbf{Average Drop}} & \multicolumn{2}{c|}{\textbf{0.58}} & \multicolumn{2}{c|}{\textbf{0.30}} & \multicolumn{2}{c|}{\textbf{0.25}} & \multicolumn{3}{c}{\textbf{Base Case}}\\
\midrule
\multicolumn{1}{c|}{\textbf{\rnnpool}} & & & & & & & & & \\
FACE-A & 0.338 & 235K & \textbf{0.635} & 213K & \textbf{0.635} & 213K & 0.626 & 232K & 232K\\
FACE-B & 0.314 & 218K & \textbf{0.320} & 218K & \textbf{0.320} & 218K & \textbf{0.320} & 218K & 165K\\
FACE-C & 0.139 & 204K & \textbf{0.548} & 188K & \textbf{0.548} & 188K & \textbf{0.548} & 188K & 185K\\
\midrule
\multicolumn{1}{c|}{\textbf{Average Drop}} & \multicolumn{2}{c|}{\textbf{0.233}} & \multicolumn{2}{c|}{\textbf{-0.003}} & \multicolumn{2}{c|}{\textbf{-0.003}} & \multicolumn{3}{c}{\textbf{Base Case}}\\
\bottomrule
\end{tabular}
\label{tab:fixedAccuracy}
\end{table*}

%% file: tables/posit_accuracy_ram.tex
\begin{table*}
\footnotesize
\caption{Performance of \toolposit on \fastgrnn, \protonn, \bonsai and \rnnpool models against computationally expensive bitwidth assignment mechanisms and the previous state-of-the-art. Accuracy is in percent, and size is in bytes. For classification-based models, the number of classes follow the dataset name. Average accuracy/MAP drop is calculated with respect to \toolposit value.}
\vspace{3mm}
\centering
\begin{tabular}{c|cc|cc|cc|ccc}
\multicolumn{1}{c|}{\textbf{Model}} & \multicolumn{2}{c}{\textbf{\previoustoolposit}} & \multicolumn{2}{|c}{\textbf{Accuracy-Based}} & \multicolumn{2}{|c}{\textbf{Size-Based}} &
\multicolumn{3}{|c}{\textbf{\toolposit}}\\ 
\multicolumn{1}{c|}{\textbf{\&}} & \multirow{2}{*}{\textbf{Accuracy}} & \multirow{2}{*}{\textbf{RAM}} & \multirow{2}{*}{\textbf{Accuracy}} & \multirow{2}{*}{\textbf{RAM}} & \multirow{2}{*}{\textbf{Accuracy}} & \multirow{2}{*}{\textbf{RAM}} & \multirow{2}{*}{\textbf{Accuracy}} & \multirow{2}{*}{\textbf{RAM}} & \multicolumn{1}{c}{\textbf{RAM}}\\ 
\multicolumn{1}{c|}{\textbf{Dataset}} & & & & & & & & & \multicolumn{1}{c}{\textbf{(OPT)}}\\
\toprule
\multicolumn{1}{c|}{\textbf{\fastgrnn}} & & & & & & & & & \\
DSA-19 & 77.00 & 448 & 77.00 & 448 & 76.91 & 448 & \textbf{77.13} & 456 & 448\\
WAKEWORD-2 & \textbf{98.40} & 160 & \textbf{98.40} & 160 & \textbf{98.40} & 160 & \textbf{98.40} & 160 & 160\\
GOOGLE-12 & 92.50 & 941 & 92.57 & 941 & 92.54 & 941 & \textbf{92.63} & 900 & 900\\
GOOGLE-30 & 77.31 & 951 & 78.64 & 951 & 78.35 & 800 & \textbf{78.65} & 902 & 900\\
HAR-2 & 90.80 & 720 & \textbf{91.18} & 720 & 91.14 & 720 & \textbf{91.18} & 720 & 720\\
HAR-6 & \textbf{91.35} & 400 & \textbf{91.35} & 400 & 91.25 & 400 & \textbf{91.35} & 400 & 400\\
USPS-10 & 94.02 & 416 & 94.02 & 416 & \textbf{94.17} & 416 & 93.82 & 416 & 416\\
MNIST-10 & 96.68 & 768 & 96.68 & 768 & \textbf{96.86} & 768 & 96.83 & 768 & 768\\
INDUSTRIAL-72 & 86.56 & 640 & \textbf{89.84} & 640 & \textbf{89.84} & 640 & \textbf{89.84} & 640 & 640\\
\midrule
\multicolumn{1}{c|}{\textbf{Average Drop}} & \multicolumn{2}{c|}{\textbf{0.58}} & \multicolumn{2}{c|}{\textbf{0.02}} & \multicolumn{2}{c|}{\textbf{0.04}} & \multicolumn{3}{c}{\textbf{Base Case}}\\
\midrule
\multicolumn{1}{c|}{\textbf{\protonn}} & & & & & & & & & \\
CIFAR-2 & 76.30 & 79 & \textbf{76.52} & 79 & 68.82 & 51 & 76.47 & 79 & 79\\
CR-2 & \textbf{73.07} & 104 & 72.85 & 104 & 66.97 & 66 & 72.80 & 104 & 104\\
CR-62 & 32.12 & 136 & \textbf{32.53} & 136 & 31.44 & 136 & 32.43 & 136 & 135\\
CURET-61 & 53.10 & 134 & \textbf{54.17} & 134 & 53.10 & 134 & \textbf{54.17} & 134 & 133\\
LETTER-26 & 82.62 & 64 & 82.62 & 64 & 82.62 & 64 & \textbf{82.70} & 64 & 63\\
MNIST-2 & 93.93 & 40 & 93.93 & 40 & 94.67 & 40 & \textbf{94.86} & 40 & 33\\
MNIST-10 & 91.31 & 41 & 91.31 & 41 & 91.37 & 41 & \textbf{91.62} & 41 & 41\\
USPS-2 & 93.17 & 40 & 93.17 & 40 & 94.02 & 40 & \textbf{94.07} & 40 & 36\\
USPS-10 & 91.83 & 810 & \textbf{92.83} & 810 & 92.43 & 810 & 92.53 & 820 & 800\\
WARD-2 & 94.41 & 54 & \textbf{94.98} & 54 & 93.79 & 36 & 94.93 & 54 & 54\\
\midrule
\multicolumn{1}{c|}{\textbf{Average Drop}} & \multicolumn{2}{c|}{\textbf{0.47}} & \multicolumn{2}{c|}{\textbf{0.17}} & \multicolumn{2}{c|}{\textbf{1.73}} & \multicolumn{3}{c}{\textbf{Base Case}}\\
\midrule
\multicolumn{1}{c|}{\textbf{\bonsai}} & & & & & & & & & \\
CIFAR-2 & 75.90 & 40 & 75.67 & 40 & \textbf{75.96} & 40 & 75.90 & 40 & 40\\
CR-2 & \textbf{74.97} & 40 & 74.66 & 40 & 74.76 & 40 & \textbf{74.97} & 40 & 40\\
CR-62 & \textbf{10.76} & 268 & 10.50 & 268 & \textbf{10.76} & 268 & \textbf{10.76} & 268 & 248\\
CURET-61 & 45.12 & 268 & 45.69 & 268 & 45.47 & 268 & \textbf{45.76} & 268 & 256\\
LETTER-26 & \textbf{65.28} & 124 & 65.26 & 118 & 65.08 & 118 & \textbf{65.28} & 124 & 104\\
MNIST-2 & 95.88 & 20 & 95.88 & 20 & \textbf{96.06} & 20 & 96.03 & 20 & 20\\
MNIST-10 & \textbf{93.49} & 50 & 93.43 & 50 & \textbf{93.49} & 50 & \textbf{93.49} & 50 & 50\\
USPS-2 & \textbf{94.92} & 60 & 94.77 & 60 & \textbf{94.92} & 60 & \textbf{94.92} & 60 & 60\\
USPS-10 & \textbf{91.83} & 48 & 91.73 & 48 & \textbf{91.83} & 48 & \textbf{91.83} & 48 & 44\\
WARD-2 & \textbf{95.18} & 20 & \textbf{95.18} & 20 & \textbf{95.18} & 20 & \textbf{95.18} & 20 & 20\\
\midrule
\multicolumn{1}{c|}{\textbf{Average Drop}} & \multicolumn{2}{c|}{\textbf{0.08}} & \multicolumn{2}{c|}{\textbf{0.14}} & \multicolumn{2}{c|}{\textbf{0.06}} & \multicolumn{3}{c}{\textbf{Base Case}}\\
\midrule
\multicolumn{1}{c|}{\textbf{\rnnpool}} & & & & & & & & & \\
FACE-A & 0.583 & 235K & 0.638 & 213K & 0.638 & 213K & \textbf{0.640} & 230K & 230K\\
FACE-B & 0.297 & 230K & 0.297 & 230K & 0.297 & 230K & \textbf{0.331} & 230K & 230K\\
FACE-C & \textbf{0.464} & 230K & \textbf{0.464} & 230K & \textbf{0.464} & 230K & \textbf{0.464} & 230K & 230K\\
\midrule
\multicolumn{1}{c|}{\textbf{Average Drop}} & \multicolumn{2}{c|}{\textbf{0.030}} & \multicolumn{2}{c|}{\textbf{0.012}} & \multicolumn{2}{c|}{\textbf{0.012}} & \multicolumn{3}{c}{\textbf{Base Case}}\\
\bottomrule
\end{tabular}
\label{tab:positAccuracy}
\end{table*}

%% file: tables/fixed_point_times.tex
\begin{table*}
\footnotesize
\caption{Inference time and compilation time of \toolfixed on \fastgrnn, \protonn, \bonsai and \rnnpool models. Inference values are in seconds, and compilation breakup is presented as a percentage of the overall compilation cost. Inference values are obtained by averaging over the final test set. For classification-based models, the number of classes follow the dataset name.}
\vspace{3mm}
\centering
\begin{tabular}{c|cc|cc|ccc}
\multicolumn{1}{c|}{\textbf{Model}} & \multicolumn{2}{c|}{\textbf{\previoustoolfixed}} & \multicolumn{5}{c}{\textbf{\toolfixed}}\\ 
\multicolumn{1}{c|}{\textbf{\&}} & \multicolumn{2}{c|}{\textbf{Time(s)}} & \multicolumn{2}{c}{\textbf{Time(s)}} & \multicolumn{3}{c}{\textbf{Compilation Breakup (\%)}}\\
\multicolumn{1}{c|}{\textbf{Dataset}} & \multicolumn{1}{c}{\textbf{Inference}} & \multicolumn{1}{c|}{\textbf{Compiler}} & \multicolumn{1}{c}{\textbf{Inference}} & \multicolumn{1}{c}{\textbf{Compiler}} & \multicolumn{1}{c}{\textbf{Stage I \& II}} & \multicolumn{1}{c}{\textbf{Stage III}} & \multicolumn{1}{c}{\textbf{\autoref{sec:DLX}}}\\
\toprule
\multicolumn{1}{c|}{\textbf{\fastgrnn}} & & & & & & &\\
DSA-19 & 1.964 & 919.77 & 2.047 & 552.6 & 94.54 & 5.45 & 0.01\\
WAKEWORD-2 & 1.167 & 429.27 & 1.124 & 242.2 & 91.52 & 8.46 & 0.02\\
GOOGLE-12 & 2.645 & 717.12 & 2.718 & 443.2 & 94.44 & 5.53 & 0.03\\
GOOGLE-30 & 3.043 & 701.65 & 3.060 & 427.2 & 91.64 & 8.32 & 0.04\\
HAR-2 & 2.690 & 288.72 & 2.633 & 143.4 & 89.96 & 10.02 & 0.02\\
HAR-6 & 3.201 & 289.90 & 2.990 & 160.9 & 81.96 & 18.01 & 0.03\\
USPS-10 & 1.094 & 183.42 & 1.172 & 106.7 & 68.22 & 31.76 & 0.02\\
MNIST-10 & 0.901 & 1765.29 & 0.903 & 1026.5 & 99.52 & 0.47 & 0.01\\
INDUSTRIAL-72 & 0.092 & 240.55 & 0.092 & 115.2 & 76.43 & 23.51 & 0.06\\
\midrule
\multicolumn{1}{c|}{\textbf{\protonn}} & & & & & & &\\
CIFAR-2 & 0.003 & 387.08 & 0.004 & 275.36 & 84.48 & 15.50 & 0.02\\
CR-2 & 0.006 & 100.07 & 0.007 & 64.72 & 72.17 & 27.78 & 0.05\\
CR-62 & 0.008 & 100.94 & 0.007 & 65.52 & 72.29 & 27.62 & 0.09\\
CURET-61 & 0.007 & 112.35 & 0.008 & 73.93 & 74.33 & 25.55 & 0.12\\
LETTER-26 & 0.010 & 94.72 & 0.010 & 60.36 & 68.45 & 31.45 & 0.10\\
MNIST-2 & 0.004 & 846.66 & 0.004 & 574.83 & 87.30 & 12.69 & 0.01\\
MNIST-10 & 0.004 & 857.41 & 0.004 & 570.29 & 87.39 & 12.60 & 0.01\\
USPS-2 & 0.003 & 116.36 & 0.003 & 69.11 & 73.25 & 26.71 & 0.04\\
USPS-10 & 0.004 & 112.52 & 0.005 & 65.39 & 79.63 & 20.30 & 0.07\\
WARD-2 & 0.004 & 136.06 & 0.005 & 92.33 & 77.50 & 22.46 & 0.04\\
\midrule
\multicolumn{1}{c|}{\textbf{\bonsai}} & & & & & & &\\
CIFAR-2 & 0.002 & 777.45 & 0.002 & 434.16 & 89.69 & 10.30 & 0.01\\
CR-2 & 0.002 & 211.18 & 0.002 & 89.89 & 77.15 & 22.82 & 0.03\\
CR-62 & 0.010 & 153.44 & 0.010 & 78.37 & 83.78 & 16.10 & 0.12\\
CURET-61 & 0.013 & 162.71 & 0.015 & 84.04 & 84.19 & 15.48 & 0.33\\
LETTER-26 & 0.004 & 127.07 & 0.005 & 71.66 & 65.60 & 34.34 & 0.06\\
MNIST-2 & 0.002 & 1469.21 & 0.003 & 823.83 & 91.21 & 8.78 & 0.01\\
MNIST-10 & 0.004 & 1284.33 & 0.005 & 774.28 & 87.55 & 12.44 & 0.01\\
USPS-2 & 0.002 & 220.63 & 0.002 & 92.87 & 77.81 & 22.15 & 0.04\\
USPS-10 & 0.003 & 172.82 & 0.003 & 99.00 & 80.54 & 19.43 & 0.03\\
WARD-2 & 0.003 & 237.05 & 0.003 & 121.28 & 82.16 & 17.82 & 0.02\\
\midrule
\multicolumn{1}{c|}{\textbf{\rnnpool}} & & & & & & &\\
FACE-A & 16.661 & 9279.46 & 16.864 & 11265.9 & 37.21 & 61.46 & 1.33\\
FACE-B & 7.880 & 2955.20 & 7.942 & 2343.09 & 55.50 & 26.61 & 17.89\\
FACE-C & 9.440 & 3060.02 & 9.140 & 2039.10 & 68.24 & 30.71 & 1.05\\
\bottomrule
\end{tabular}
\label{tab:fixedPointTimes}
\end{table*}

%% file: tables/fixed_point_times_uno.tex
\begin{table*}
\footnotesize
\caption{Inference time of \toolfixed on Arduino Uno board for \fastgrnn, \protonn and \bonsai models. Inference values are in seconds, and are obtained by averaging over the final test set. The number of classes follow the dataset name.}
\vspace{3mm}
\centering
\begin{tabular}{c|c}
\multicolumn{1}{c|}{\textbf{Model}} & \multicolumn{1}{c}{\textbf{\toolfixed}}\\ 
\multicolumn{1}{c|}{\textbf{\&}} & \multicolumn{1}{c}{\textbf{Time(s)}}\\
\multicolumn{1}{c|}{\textbf{Dataset}} & \multicolumn{1}{c}{\textbf{Inference}}\\
\toprule
\multicolumn{1}{c|}{\textbf{\fastgrnn}} &\\
DSA-19 & 56.18\\
WAKEWORD-2 & 27.61\\
GOOGLE-12 & 75.18\\
GOOGLE-30 & 98.72\\
HAR-2 & 93.48\\
HAR-6 & 94.38\\
USPS-10 & 2.70\\
MNIST-10 & 25.89\\
INDUSTRIAL-72 & 1.70\\
\midrule
\multicolumn{1}{c|}{\textbf{\protonn}} &\\
CIFAR-2 & 0.07\\
CR-2 & 0.13\\
CR-62 & 0.27\\
CURET-61 & 0.07\\
LETTER-26 & 0.20\\
MNIST-2 & 0.10\\
MNIST-10 & 0.11\\
USPS-2 & 0.05\\
USPS-10 & 0.11\\
WARD-2 & 0.11\\
\midrule
\multicolumn{1}{c|}{\textbf{\bonsai}} &\\
CIFAR-2 & 0.08\\
CR-2 & 0.08\\
CR-62 & 0.31\\
CURET-61 & 0.36\\
LETTER-26 & 0.07\\
MNIST-2 & 0.08\\
MNIST-10 & 0.09\\
USPS-2 & 0.08\\
USPS-10 & 0.04\\
WARD-2 & 0.10\\
\bottomrule
\end{tabular}
\label{tab:fixedPointTimesUno}
\end{table*}